%% file: main.tex
\documentclass[10pt,journal,compsoc]{IEEEtran}
\ifCLASSOPTIONcompsoc
\usepackage[nocompress]{cite}
\else
\usepackage{cite}
\fi
\ifCLASSINFOpdf
 \usepackage[pdftex]{graphicx}
\else
\usepackage[dvips]{graphicx}
\DeclareGraphicsExtensions{.eps}
\fi
\usepackage{amsmath}
\usepackage{amssymb}
\usepackage{amsthm}
\usepackage{array}
\usepackage{url}
\usepackage[linesnumbered,ruled]{algorithm2e}

\theoremstyle{definition}

\let\oldnl\nl
\newcommand{\nonl}{\renewcommand{\nl}{\let\nl\oldnl}}

\usepackage{multirow}
\usepackage[font=scriptsize]{subfig}
\usepackage{epstopdf}
\usepackage{subfiles}
\usepackage{makecell}
\usepackage{epstopdf}
\usepackage{subfiles}
\usepackage{makecell}
\usepackage{soul}
\soulregister{\ref}7
\soulregister{\cite}7
\usepackage{color, xcolor}

\usepackage{pifont}

\usepackage{verbatim}

\soulregister\cite7 
\soulregister\eqref7
\usepackage{booktabs}
\usepackage{colortbl}

\usepackage{algorithmic}
\usepackage{textcomp}
\usepackage{stfloats}
\usepackage{booktabs}
\usepackage{tabularx}
\usepackage{stmaryrd}
\usepackage{enumitem}
\usepackage{wrapfig}
\usepackage{threeparttable}
\usepackage{ulem}

\definecolor{improvegreen}{RGB}{0, 128, 0} 
\definecolor{improveblue}{RGB}{0, 0, 255} 
\definecolor{improvered}{RGB}{0, 0, 0} 

\setlist[itemize]{leftmargin=10pt}

\begin{document}

\title{AtomThink: Multimodal Slow Thinking with Atomic Step Reasoning}

\author{Kun Xiang\IEEEauthorrefmark{1}, 
Zhili Liu\IEEEauthorrefmark{1},
Terry Jingchen Zhang, 
Yinya Huang, 
Yunshuang Nie, 
Kaixin Cai,  \\
Yiyang Yin,
Runhui Huang, 
Hanhui Li\IEEEauthorrefmark{2}, 
Yihan Zeng, 
Yu-Jie Yuan, 
Jianhua Han, \\
Lanqing Hong, 
Hang Xu, 
Xiaodan Liang\IEEEauthorrefmark{2}

\IEEEcompsocitemizethanks{
	\IEEEcompsocthanksitem 
	\IEEEauthorrefmark{1}These two authors contribute equally to this work.
        \vspace{0.2cm}
	\IEEEcompsocthanksitem 
	\IEEEauthorrefmark{2}Xiaodan Liang and Hanhui Li are the corresponding authors.
        \vspace{0.2cm}
        \IEEEcompsocthanksitem Kun Xiang, Yunshuang Nie, Kaixin Cai, Yiyang Yin and Hanhui Li are with Shenzhen Campus of Sun Yat-sen University, Shenzhen, China. 
        \protect\\ E-mail: \{xiangk@mail2.sysu.edu.cn\}
        \IEEEcompsocthanksitem Terry Jingchen Zhang and Yinya Huang are with ETH Zurich, Zurich, Switzerland. 
        \IEEEcompsocthanksitem Zhili Liu is with the Hong Kong University of Science and Technology, Hong Kong, China.
        \IEEEcompsocthanksitem Runhui Huang is with the University of Hong Kong, Hong Kong, China.
        \IEEEcompsocthanksitem Yihan Zeng, Yu-Jie Yuan, Jianhua Han, Lanqing Hong and Hang Xu are with Noah's Ark Lab, Shanghai, China. 
        \IEEEcompsocthanksitem Jianhua Han and Hang Xu are with Yinwang Intelligent Technology Co., Ltd., Shanghai, China. 
        \IEEEcompsocthanksitem Xiaodan Liang is with Shenzhen Campus of Sun Yat-sen University, Shenzhen, China, Peng Cheng Laboratory, Guangdong Key Laboratory of Big Data Analysis and Processing, Guangzhou, 510006, China. 
        \protect\\E-mail: \{liangxd9@mail.sysu.edu.cn\}
	}
}


\maketitle
\input{sec/0_abstract}    
\input{sec/1_intro}

\input{sec/2_related_work}
\input{sec/3_method}
\input{sec/4_exp}

\input{sec/5_limitation}
\input{sec/6_conclusion}

\section*{Acknowledgments}
This work is supported by Scientific Research Innovation Capability Support Project for Young Faculty (No.ZYGXQNJSKYCXNLZCXM-I28), National Natural Science Foundation of China (NSFC) under Grants No.62476293 and No.62372482, and General Embodied AI Center of Sun Yat-sen University.

{
    \small
    \bibliographystyle{IEEEtran}
    \bibliography{main}
}

\begin{IEEEbiography}
[{\includegraphics[width=1in,height=1.25in, clip,keepaspectratio]{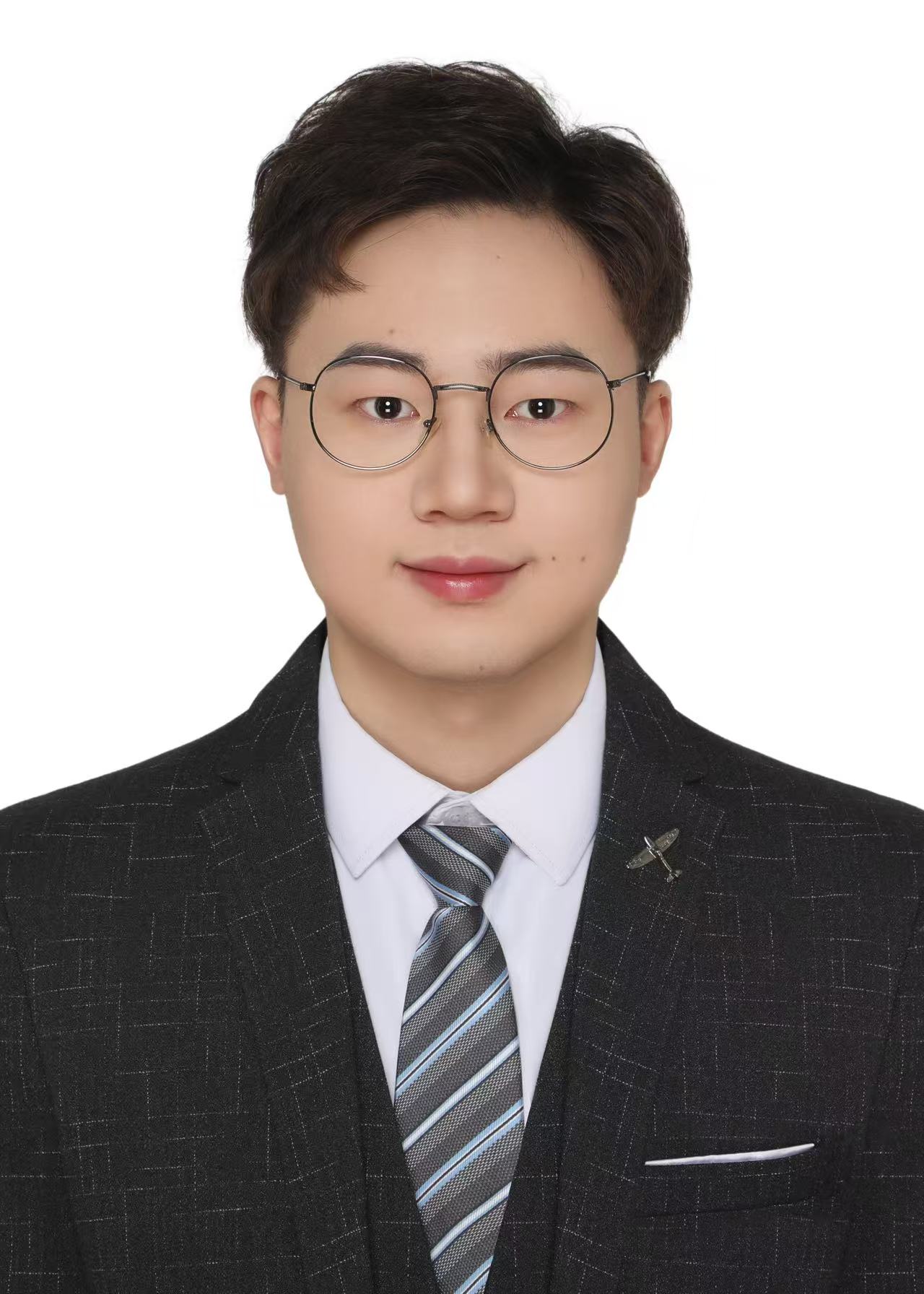}}]{Kun Xiang} is currently a PhD student at HCP-I2 Lab in Sun Yat-sen University advised by Prof. Xiaodan Liang. He has received his B.S. degree and M.S. degree from School of Intelligent Systems Engineering in Sun Yat-sen University, China, in 2021 and 2024, respectively. He is interested in generalizable multimodal AI systems and high order reasoning capability in MLLMs.
\end{IEEEbiography}

\begin{IEEEbiography}
[{\includegraphics[width=1in,height=1.25in, clip,keepaspectratio]{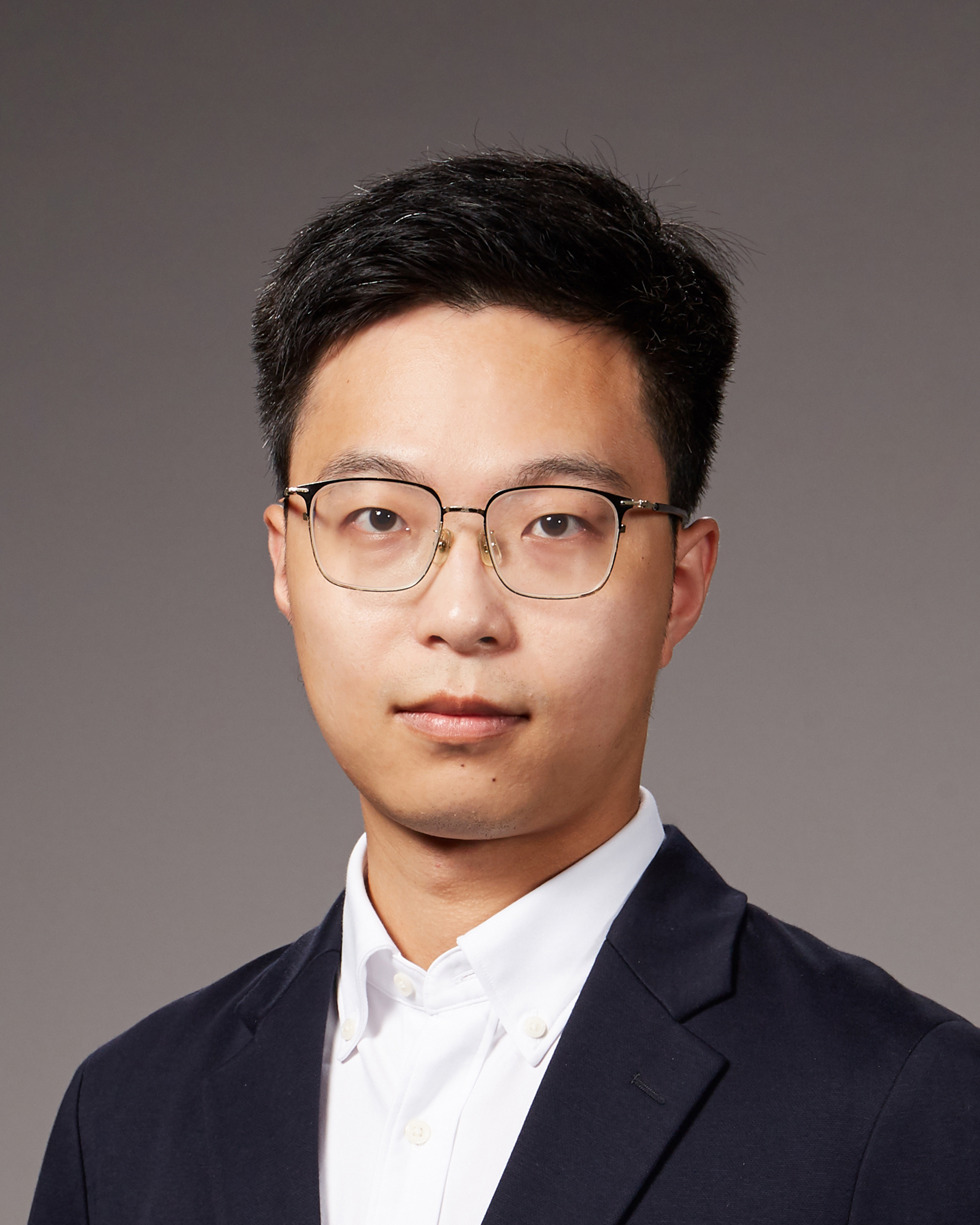}}]{Zhili Liu} is currently a Ph.D. candidate at the Hong Kong University of Science and Technology, working under the supervision of Prof. James Kwok. He received his Bachelor’s degree in Software Engineering from Tongji University in 2018, and his Master’s degree in Computer Science from The Chinese University of Hong Kong in 2019. His research interests include multimodal large language models and mixture-of-experts systems.
\end{IEEEbiography}

\begin{IEEEbiography}
[{\includegraphics[width=1in,height=1.25in, clip,keepaspectratio]{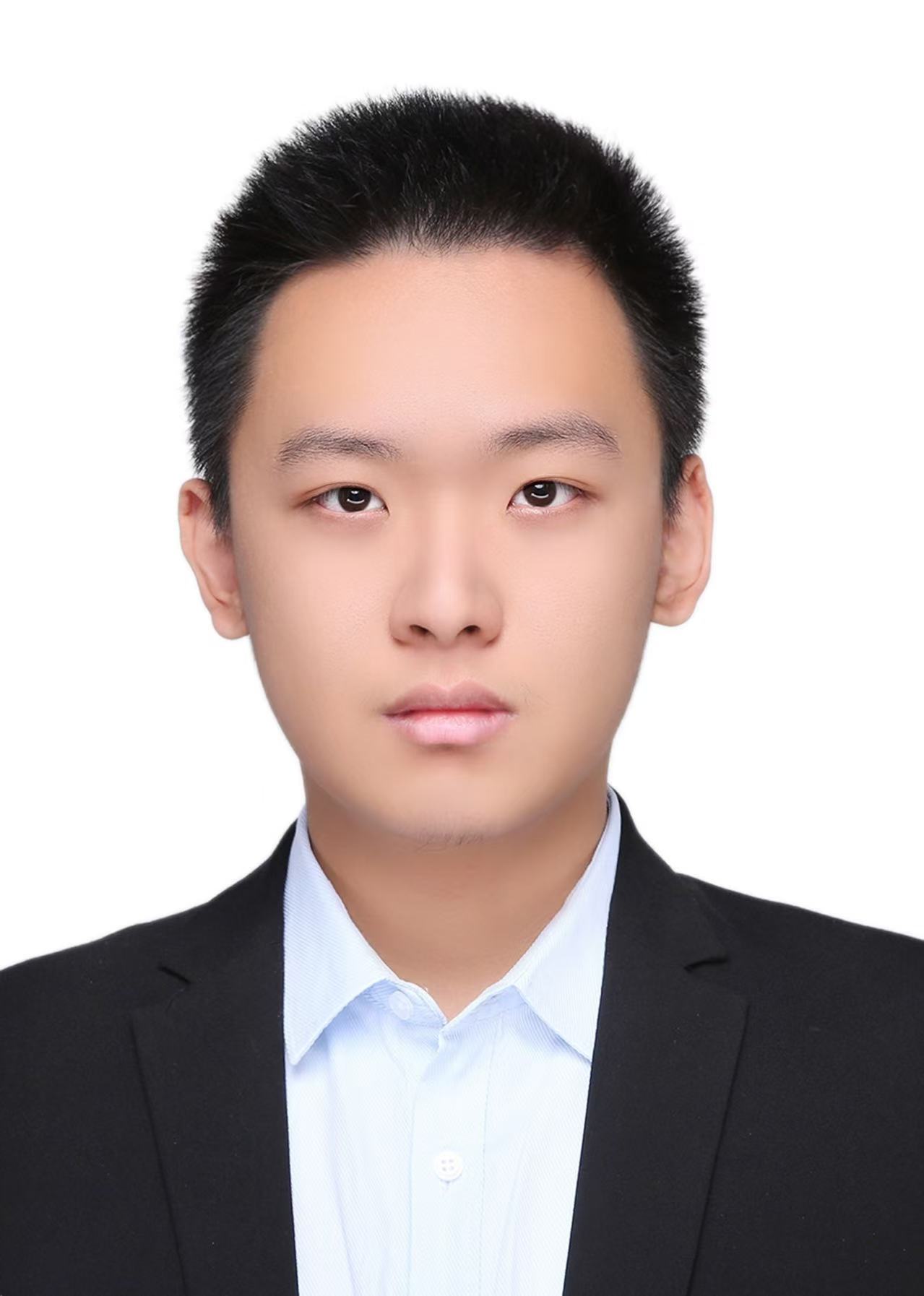}}]{Terry Jingchen Zhang} is currently a student in the interdisciplinary science program directed by Prof. Jeremy Richardson at ETH Zurich, Zurich, Switzerland. His research interest include AI-driven scientific discovery, AI safety and alignment science.
\end{IEEEbiography}

\begin{IEEEbiography}
[{\includegraphics[width=1in,height=1.25in, clip,keepaspectratio]{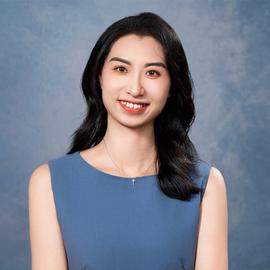}}]{Yinya Huang} is a Postdoc Fellow at ETH AI Center, ETH Zürich, working with Prof. Mrinmaya Sachan and Prof. Elliot Ash. She received her Ph.D. Degree in Computer Science from Sun Yat-sen University, advised by Prof. Xiaodan Liang and Prof. Liang Lin. Her research focuses on models’ system 2 thinking, especially incorporating formal systems and methods into LLM methods for mathematical, logical, and causal reasoning. 
\end{IEEEbiography}

\begin{IEEEbiography}
[{\includegraphics[width=1in,height=1.25in, clip,keepaspectratio]{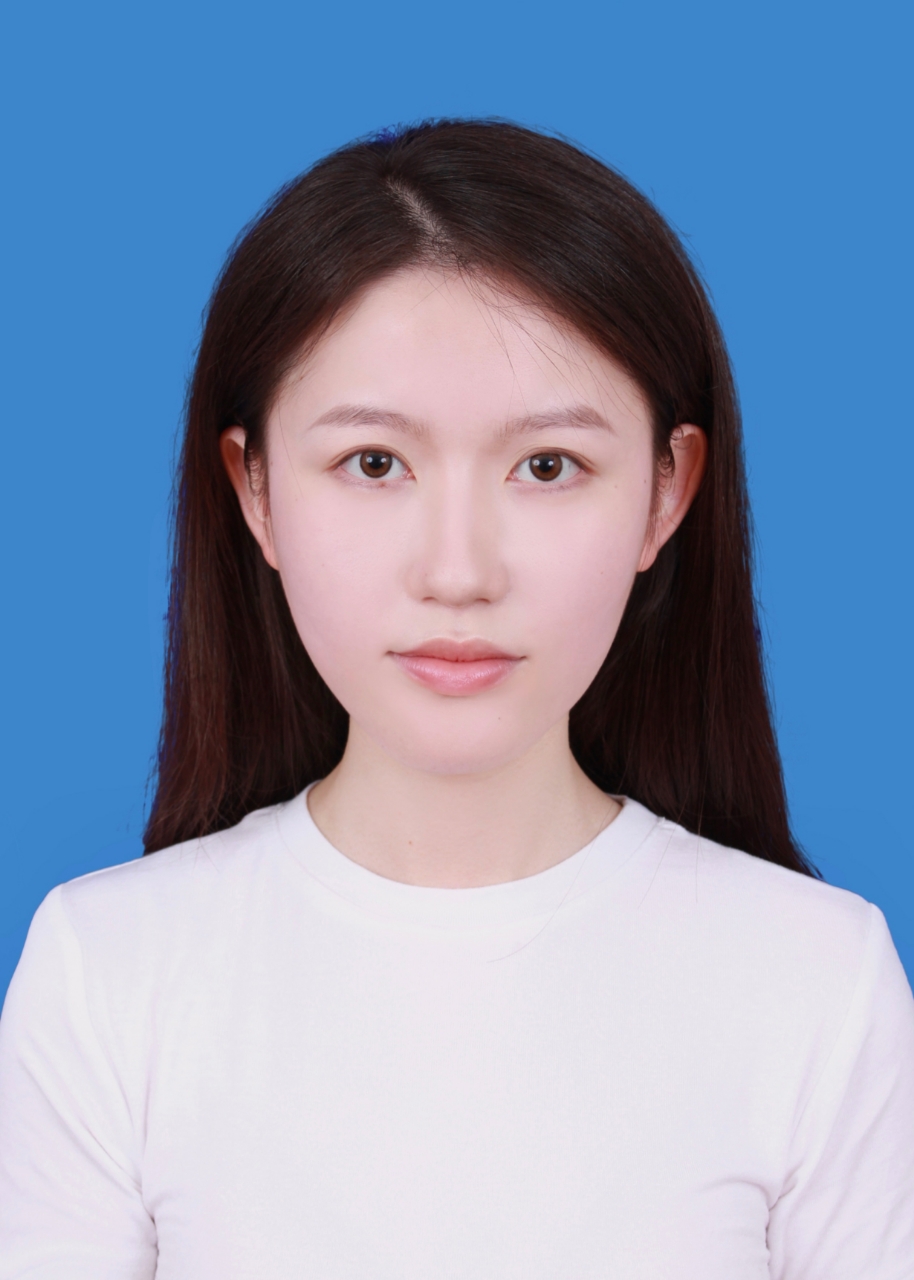}}]{Yunshuang Nie} received the B.E. degree in Sun Yat-sen University, Shenzhen, China, in 2023. She is currently working toward the M.E. in the school of intelligent systems engineering of Sun Yat-sen University. Her current research interests are multi-modality learning and embodied AI.
\end{IEEEbiography}

\begin{IEEEbiography}
[{\includegraphics[width=1in,height=1.25in, clip,keepaspectratio]{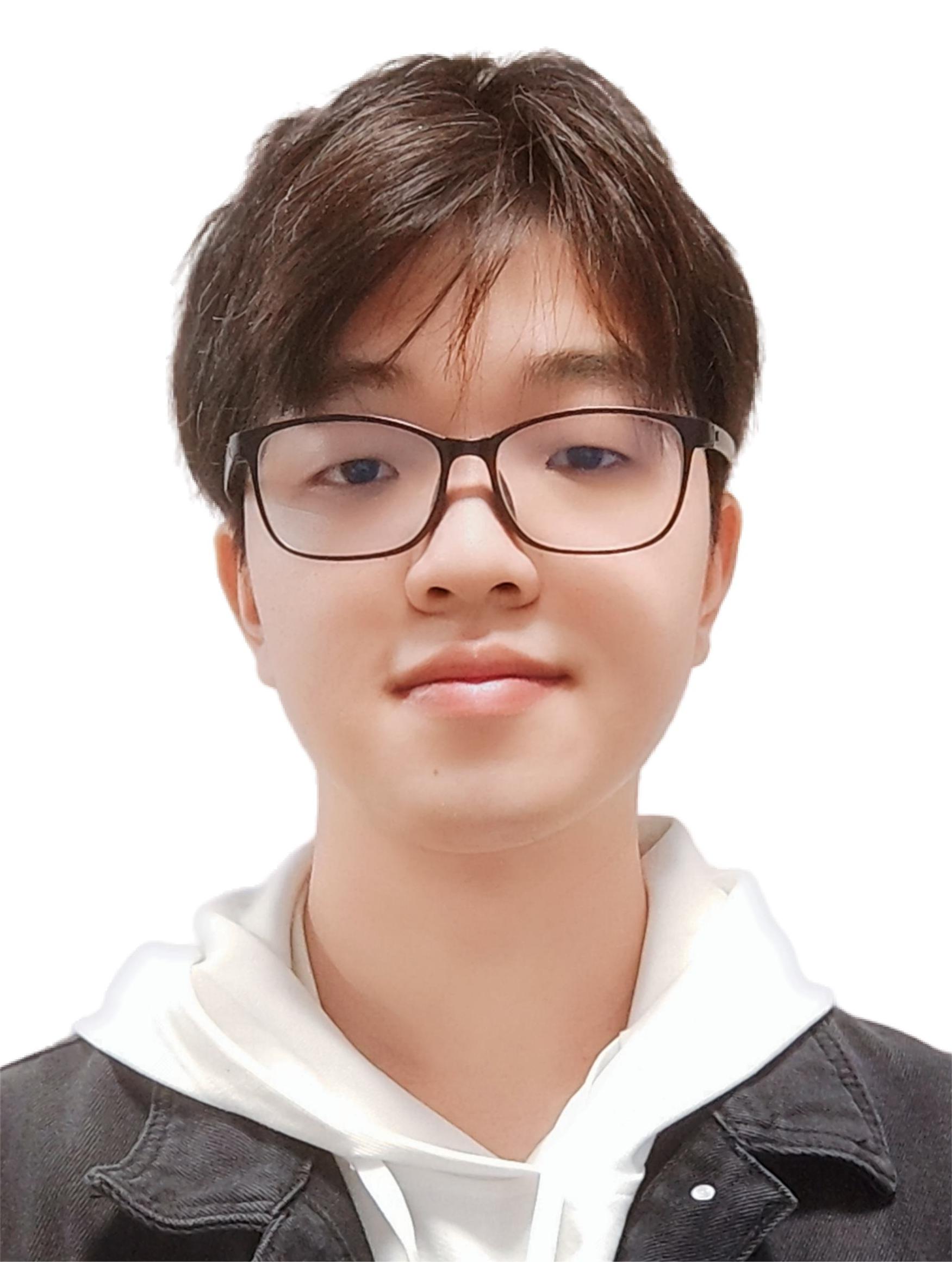}}]{Kaixin Cai} is currently a postgraduate student at the School of Intelligent Engineering, Sun Yat-Sen University, Shenzhen, China. His research interests include Segmentation, multimodal learning, and image editing.
\end{IEEEbiography}

\begin{IEEEbiography}
[{\includegraphics[width=1in,height=1.25in, clip,keepaspectratio]{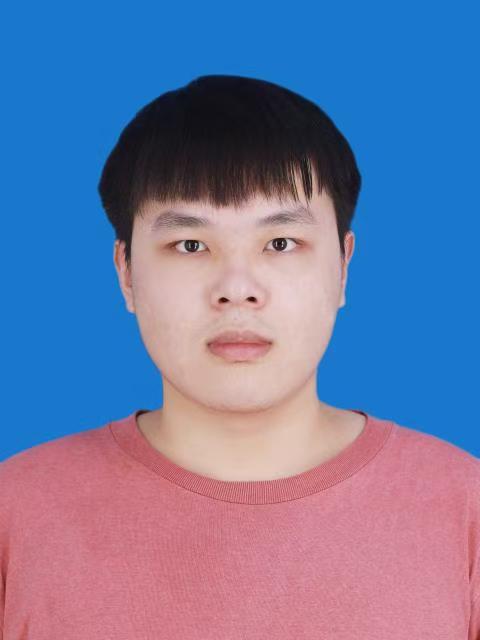}}]{Yiyang Yin} received the B.E. degree from Sun Yat-sen University. He is currently a MA.Eng with the School of Intelligent Systems Engineering, Shenzhen Campus of Sun Yat-sen University. His research focuses on Multi-model Large language models and image segmentation in computer vision, particularly open vocabulary segmentation. 
\end{IEEEbiography}

\begin{IEEEbiography}
[{\includegraphics[width=1in,height=1.25in, clip,keepaspectratio]{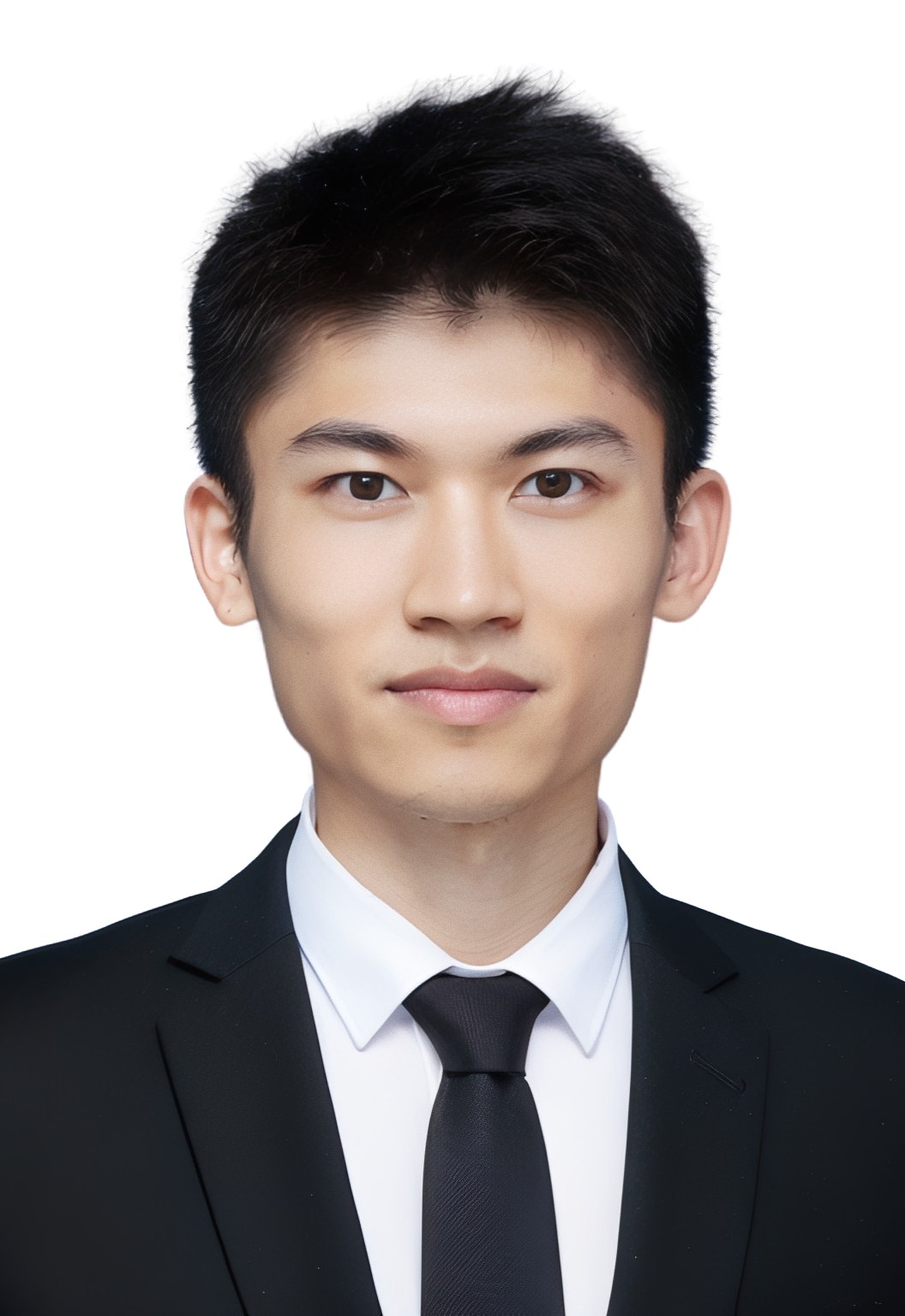}}]{Runhui Huang} received his B.S. degree and M.S. degree from Sun Yat-sen University, China, in 2021 and 2024, respectively. He is currently a Ph.D. student at the University of Hong Kong, supervised by Prof. Hengshuang Zhao. His research interests include vision-language pretraining, multimodal large language models (MLLMs), and unified multimodal understanding and generation models.
\end{IEEEbiography}

\begin{IEEEbiography}
[{\includegraphics[width=1in,height=1.25in, clip,keepaspectratio]{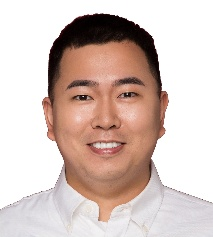}}]{Hanhui Li} received the B.S. degree in computer science and technology and the Ph.D. degreein computer software and theory from Sun Yat-sen University, Guangzhou, China, in 2012 and 2018, respectively. He is currently a Research Associate Professor with Sun Yat-sen University, Shenzhen Campus. Before that, he was a Research Fellow with Nanyang Technological University, Singapore, from 2019 to 2021. His research interests include visual media analysis and reasoning.
\end{IEEEbiography}

\begin{IEEEbiography}
[{\includegraphics[width=1in,height=1.25in, clip,keepaspectratio]{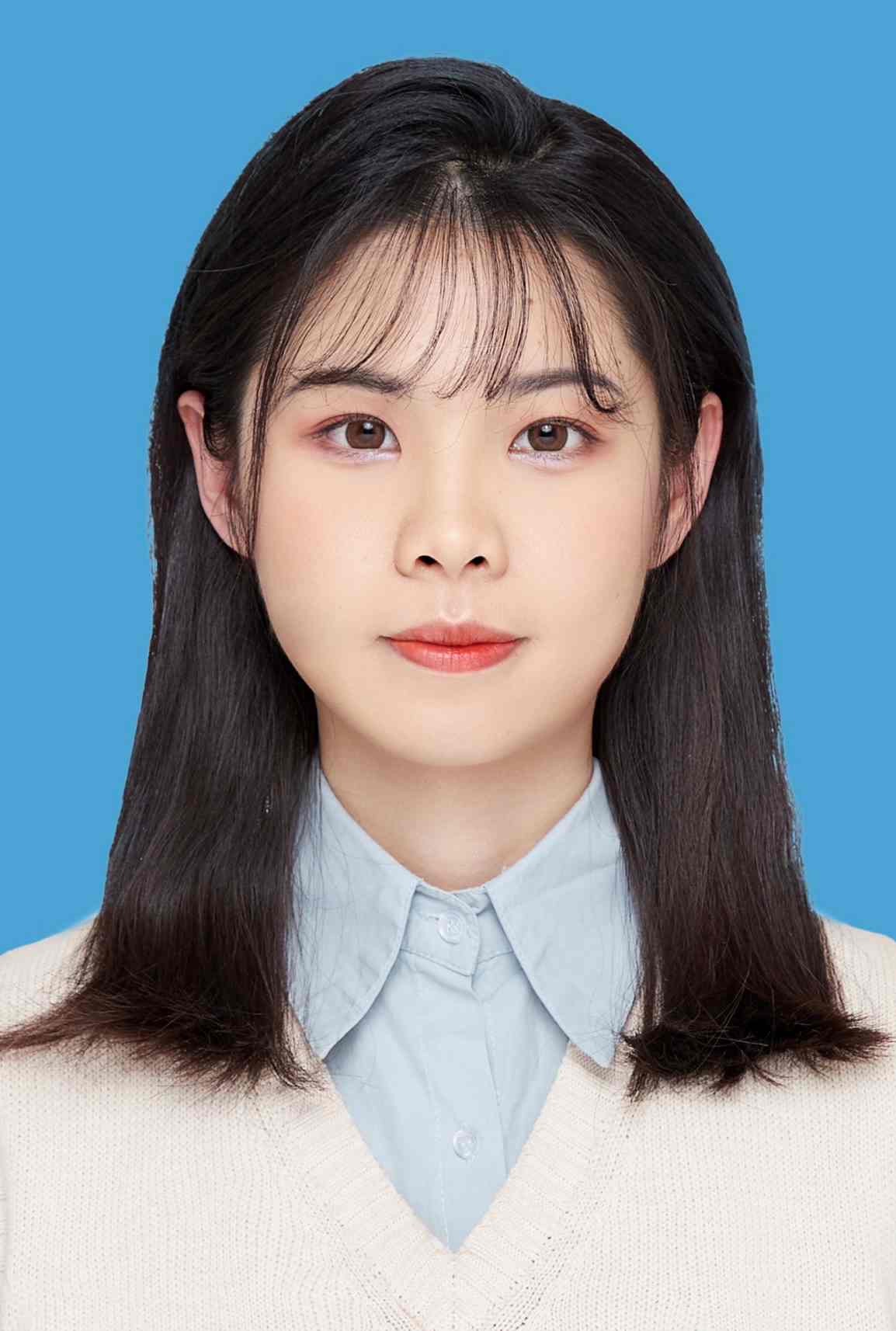}}]{Yihan Zeng} received her B.Eng. degree (2019) and M.S. degree (2022) from Shanghai Jiao Tong University. Since 2022, she has been serving as an Algorithm Engineer at Huawei Technologies Co., Ltd. Her research primarily focuses on computer vision, with a particular emphasis on 3D object detection and tracking, open-vocabulary perception, 3D generation and Multimodal Large Language Models.
\end{IEEEbiography}

\begin{IEEEbiography}
[{\includegraphics[width=1in,height=1.25in, clip,keepaspectratio]{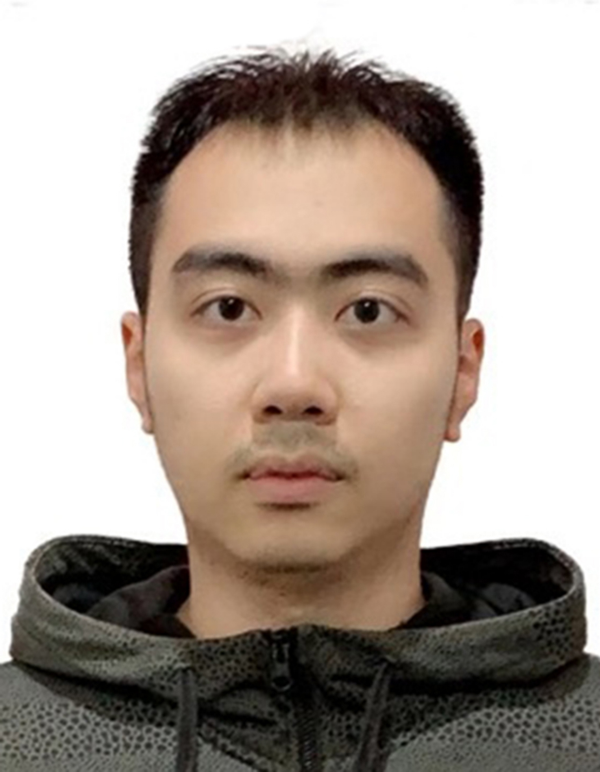}}]{Yu-Jie Yuan} received the Ph.D. degree in the Institute of Computing Technology, Chinese Academy of Sciences. His research interests include 3D learning and multimodal large language models. 
\end{IEEEbiography}

\begin{IEEEbiography}
[{\includegraphics[width=1in,height=1.25in, clip,keepaspectratio]{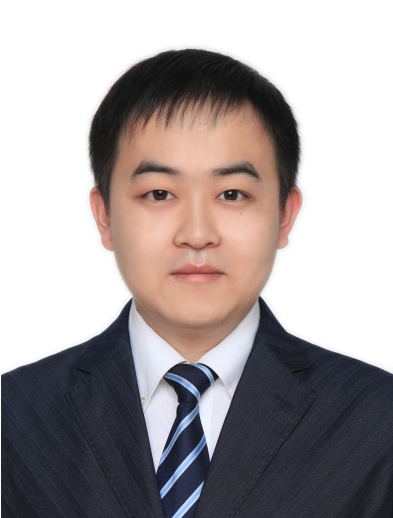}}]{Jianhua Han} received the Bachelor Degree in 2016 and Master Degree in 2019 from Shanghai Jiao Tong University, China. He is currently a re-searcher with the Noahs Ark Laboratory, Huawei Technologies Co ., Ltd. His research interests lie primarily in deep learning and computer vision.
\end{IEEEbiography}

\begin{IEEEbiography}
[{\includegraphics[width=1in,height=1.25in, clip,keepaspectratio]{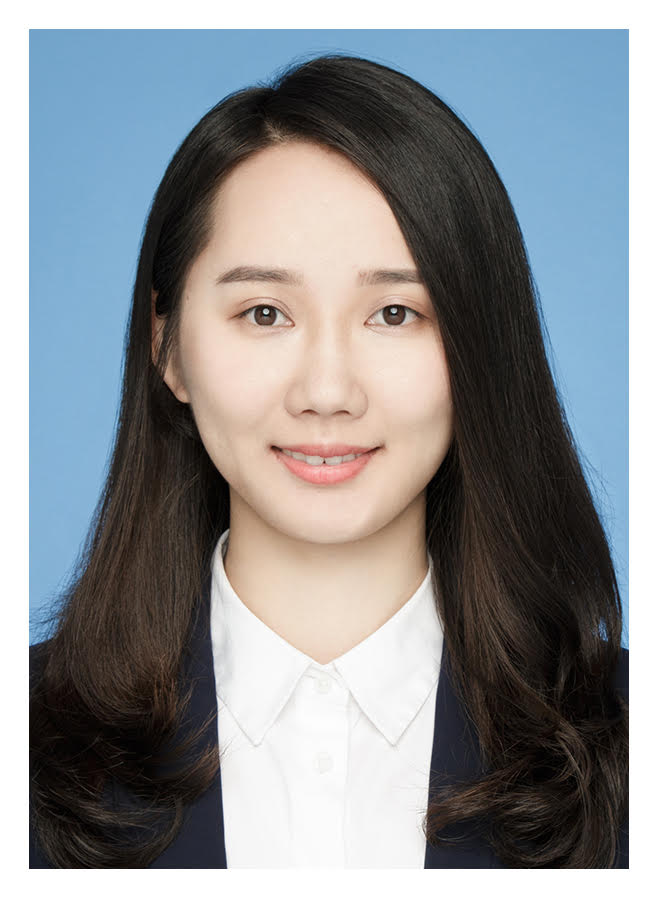}}]{Lanqing Hong} holds a Ph.D. from the National University of Singapore. Her research focuses on multimodal large models and generative AI, with an emphasis on understanding the strengths and weaknesses of existing large models, exploring their boundaries, and proposing efficient next-generation models and algorithms. She has published over 30 papers at top AI conferences, with more than 3,000 citations on Google Scholar. Dr. Hong has served as a reviewer for prestigious conferences such as NeurIPS, ICLR, and CVPR, and she is the Area Chair for IJCAI 2025 and the Industrial Chair for 3DV 2025.
\end{IEEEbiography}

\begin{IEEEbiography}
[{\includegraphics[width=1in,height=1.25in, clip,keepaspectratio]{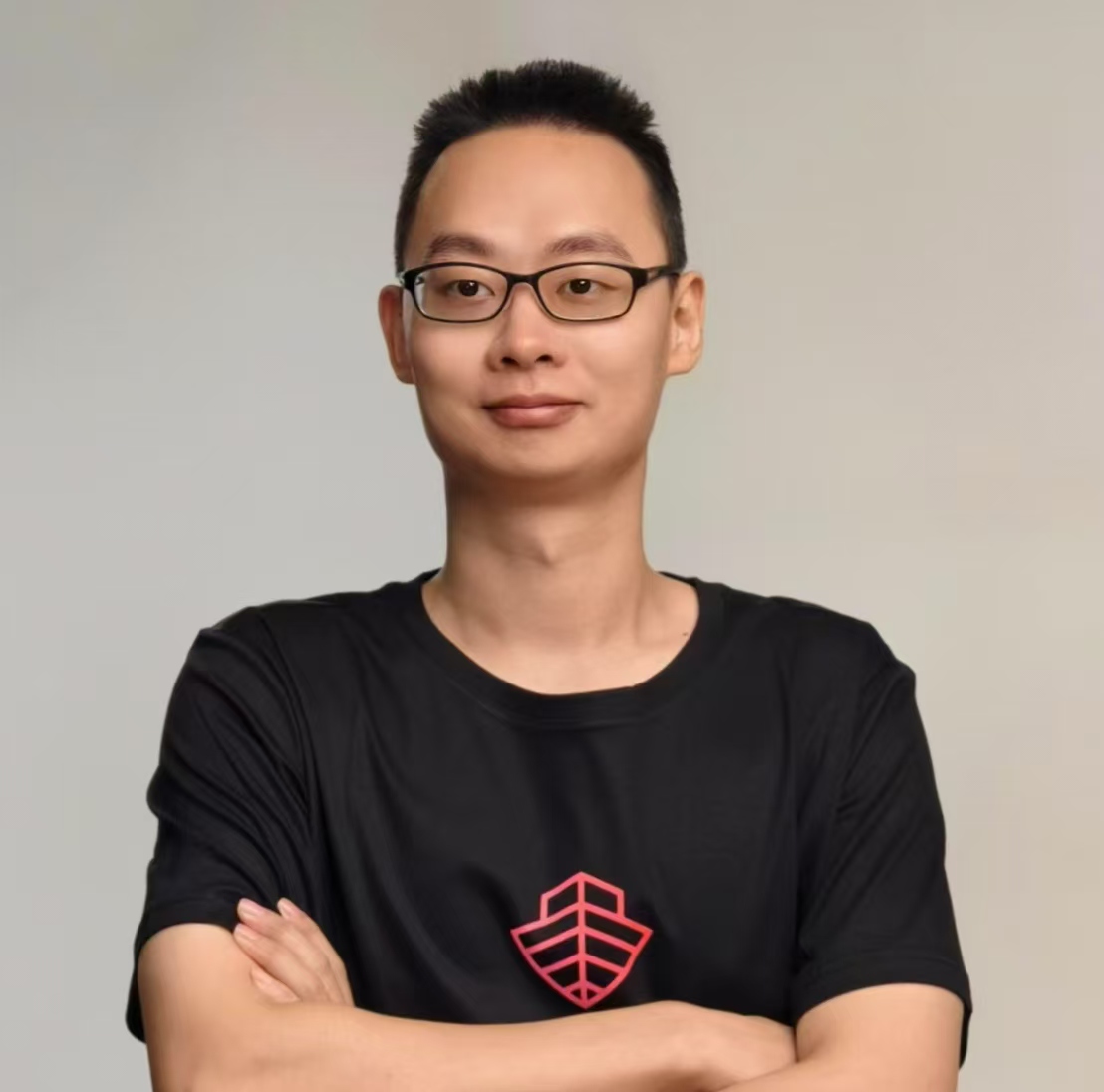}}]{Hang Xu} is currently a senior CV researcher at Huawei Noah's Ark Lab. He received his BSc from Fudan University and his Ph.D. from the University of Hong Kong. His research interests include multimodal large language models, autonomous driving, object detection, and AutoML. He has published over 100 papers at top AI conferences such as NeurIPS, CVPR, ICCV, AAAI.
\end{IEEEbiography}

\begin{IEEEbiography}
[{\includegraphics[width=1in,height=1.25in, clip,keepaspectratio]{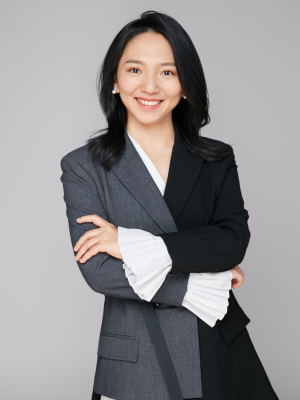}}]{Xiaodan Liang} is an Associate Professor in the Department of Computer Vision at the Mohamed bin Zayed University of Artificial Intelligence (MBZUAI), and a joint Professor at Sun Yat-sen Univerisity, China. She was a Project  Scientist at Carnegie Mellon University, working with Prof. Eric Xing. She has published over 120 cutting-edge papers on visual-language understanding and generation, and its application on embodied AI, which have appeared in the most prestigious journals and conferences in the field, with Google Citation 30000+. She serves as regular Area Chairs of ICCV, CVPR, NeurIPS, ICML, ICLR and AAAI regularly, and Tutorial Chair of CVPR 2021, Ombud chair of CVPR 2023, Local chairs of ICCV 2029. She has been awarded ACM China and CCF Best Doctoral Dissertation Award and Alibaba DAMO Academy Young Fellow. Her research has been applied in the key products in several renowned AI companies such as Deepseek, Lenovo, ByteDance and Tencent Inc.
\end{IEEEbiography}


\clearpage

\input{sec/X_suppl}

\end{document}

%% file: sec/0_abstract.tex
\begin{abstract}
In this paper, we address the challenging task of multimodal reasoning by incorporating the notion of ``slow thinking'' into multimodal large language models (MLLMs). Our core idea is that models can learn to adaptively use different levels of reasoning to tackle questions of varying complexity. We propose a novel paradigm of Self-structured Chain of Thought (SCoT), which consists of minimal semantic atomic steps. Unlike existing methods that rely on structured templates or free-form paradigms, our method  not only generates flexible CoT structures for various complex tasks but also mitigates the phenomenon of overthinking for easier tasks. To introduce structured reasoning into visual cognition, we  design a novel AtomThink framework with four key modules: (i) a data engine to generate high-quality multimodal reasoning paths; (ii) a supervised fine-tuning (SFT) process with serialized inference data; (iii) a policy-guided multi-turn inference method; and (iv) an atomic capability metric to evaluate the single-step utilization rate. Extensive experiments demonstrate that the proposed AtomThink significantly improves the performance of baseline MLLMs, achieving more than 10\% average accuracy gains on MathVista and MathVerse. Compared to state-of-the-art structured CoT approaches, our method not only achieves higher accuracy but also improves data utilization by 5 $\times$ and boosts inference efficiency by 85.3\%. Our code is publicly available at https://github.com/Kun-Xiang/AtomThink.
\end{abstract}

%% file: sec/1_intro.tex
\section{Introduction}
\label{sec:intro}

\begin{figure*}[t]
    \centering
\includegraphics[width=1.0\textwidth]{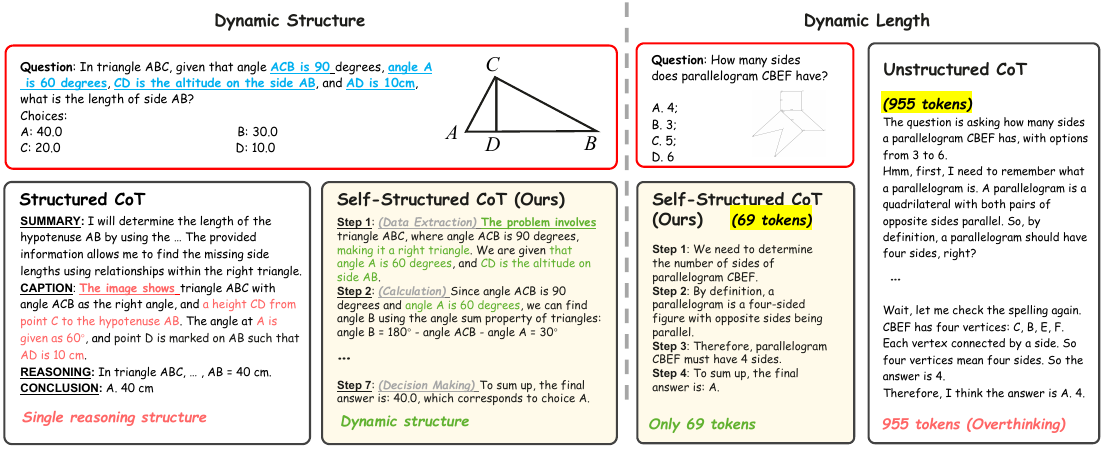} 
    \caption{Comparison with Structured-CoT (LLaVA-CoT) and Unstructured CoT (Qwen-2.5VL-72B) methods. \textcolor{improvered}{Structured-CoT enforces fixed templates (e.g., mandatory image captioning), while unstructured CoT exhibits redundancy in simple problems. Our method adapts both structure and length dynamically, skipping unnecessary steps and improving efficiency.} }
    \label{fig:intro}
\end{figure*}

Chain-of-Thought (CoT) reasoning~\cite{wei2022chain} constitutes a pivotal paradigm that substantially enhances the capacity of Large Language Models (LLMs) to address complex scientific problems. This methodology facilitates emergent intermediate reasoning steps within LLMs, exemplified by significant advances in frontier Large Reasoning Models (LRMs)~\cite{openai2024_o1,guo2025deepseek}. These models solve intricate problems through \textcolor{improvered}{extensive} reasoning chains often conceptualized as "slow thinking"~\cite{slowthinking}.

Recent research has focused on elucidating internal reasoning mechanisms in frontier LRMs~\cite{yao2024tree,gao2024interpretable,qin2024o1,wang2024openr}. Approaches such as LLaVA-CoT~\cite{llavacot} and LlamaV-o1~\cite{llamavo1} implement \textbf{Structured CoT} through fixed modules driven by manually defined templates, constraining reasoning diversity in multimodal contexts. In contrast, models including OpenAI-o1~\cite{openai2024_o1} and DeepSeek-R1~\cite{guo2025deepseek} employ \textbf{Unstructured CoT}, which eliminates predefined frameworks to autonomously generate emergent free-form reasoning chains via iterative refinement. Although Unstructured CoT better approximates human cognition and demonstrates superior generalization, recent investigations~\cite{chen2024not,wang2025thoughts} reveal these slow-thinking models suffer from inefficient token utilization and overthinking tendencies when processing simpler problems. As Figure~\ref{fig:intro} illustrates, both paradigms exhibit significant limitations. Consequently, we establish two fundamental principles: \textbf{different problems demand distinct reasoning capabilities, and reasoning chain complexity should align with problem difficulty for optimal performance}.

To dynamically generate appropriate reasoning structures for problems with diverse complexity, we introduce a novel \textcolor{improvered}{reasoning} paradigm of \textbf{Self-structured Chain-of-Thought (SCoT)}, \textcolor{improvered}{which is autonomously generated and length-controlled by the model, decomposing complex reasoning processes into atomic, verifiable steps.}  To activate the model's self-structured reasoning abilities in multimodal tasks, we further develop a full-process slow-thinking framework called \textbf{AtomThink}. \textcolor{improvered}{As a full-pipeline framework,} it comprises four key components: a data engine, supervised fine-tuning, policy search and atomic capability evaluation. To begin with, a data annotation engine with novel prompting and bad-case filtering strategies is used to create a novel multimodal long CoT dataset. We propose a dataset called AMATH, including 20k high-level mathematical problems with 124k atomic step annotations. Furthermore, our atomic step finetuning strategy applies step-level masking to the training set, forcing our models to learn individual inference steps. During the inference phase, the model is not only capable of spontaneously generating CoT in fast-thinking mode, but also continuously improve with process supervision models and step search mechanisms. Lastly, we propose an atomic capability evaluation metric based on reasoning behavior clustering and step utilization calculation, which quantitatively shows the model's capability in utilizing individual atomic steps. 

To validate the effectiveness of our method, we conduct extensive experiments on public benchmarks. 
\textcolor{improvered}{Compared to baseline models, we achieve an average accuracy improvement of \textbf{16.6\%} across four wide-adopted benchmarks for mathematical reasoning. The improvement in cross-domain generalization on general and scientific tasks is also impressive, such as a \textbf{25.3\%} improvement on TextVQA~\cite{TextVQA} and a \textbf{20.2\%} improvement on ScienceQA~\cite{ScienceQA}.}
Our approach achieves \textbf{5 times} higher data utilization than previous frontier approach LLaVA-CoT while maintaining superior performance, and we offer enhanced inference efficiency by over \textbf{80\%}. To advance multimodal reasoning research, we further provide a comprehensive fine-grained analysis of required reasoning capabilities in visual understanding models.

Our primary contributions are as follows:
\begin{itemize}
  \item We introduce \textbf{Self-structured Chain-of-Thought} as a new thinking paradigm to decompose any reasoning process into atomic steps. It eliminates the need for constructing structured thought templates and achieves significant improvements in both data utilization and inference efficiency.
  \item We offer a comprehensive \textbf{AtomThink} framework including plug-and-play modules for data annotation, atomic fine-tuning, multi-turn inference and capability evaluation, is designed to improve the reasoning ability of MLLMs.
  \item \textcolor{improvered}{We conduct extensive experiments across 11 benchmarks (mathematical, scientific and general tasks) with models of different scales. Results demonstrate consistent improvements on both In-Distribution and Out-of-Distribution tasks.} Additionally, we present a fine-grained analysis of comprehension capabilities distribution.
\end{itemize}

%% file: sec/2_related_work.tex
\section{Related Work}
\label{sec:formatting}
\subsection{Chain of Thought in Multimodal Reasoning Tasks}
Complex reasoning tasks such as mathematical computation have long been challenging for MLLMs~\cite{yin2023survey,liu2023mathematical}. \textcolor{improvered}{Prior work has addressed this challenge by encouraging models to generate Chain of Thought (CoT) reasoning to enhance their reasoning capabilities} \cite{wei2022chain,wang2022self}. \textcolor{improvered}{These methods modify} the input distribution to generate unstructured reasoning \textcolor{improvered}{paths} without finetuning parameters.
Recently, OpenAI o1 and DeepSeek R1 have demonstrated the scalability of unstructured CoT  through Reinforcement Learning. However, \textcolor{improvered}{these models} still suffer from overthinking and excessive computational consumption. Other studies have guided multimodal models to generate structured CoT by providing manually designed templates~\cite{llavacot,llamavo1}. \textcolor{improvered}{While} these models incorporate visual semantic information into the reasoning process, their fixed steps constrain the diversity of reasoning actions \textcolor{improvered}{and limit} their generalization ability on complex problems.
\subsection{Long CoT Annotation for Multimodal Data}
The introduction of slow thinking relies heavily on the availability of high-quality step-level annotations. Lightman et al. \cite{lightman2023let} constructed a process supervision dataset \textcolor{improvered}{with} extensive human annotations \textcolor{improvered}{that has} been widely used for mathematical reasoning. Recent advancements have focused on automating the data acquisition process \textcolor{improvered}{by allowing} models to generate their own CoTs. Techniques like Quiet-STaR~\cite{zelikman2024quiet} have demonstrated how self-generated reasoning can enhance model performance without requiring manual labels. \textcolor{improvered}{Some} methods based on Monte Carlo estimation have automated the  data collection \textcolor{improvered}{process but introduce} additional computational cost~\cite{Math-shepherd,OmegaPRM}. In \textcolor{improvered}{the multimodal domain}, MAVIS~\cite{zhang2024mavis} \textcolor{improvered}{is a} dataset consisting of 834k visual math problems annotated with short CoT \textcolor{improvered}{that has been proposed}. Other studies have distilled reasoning processes from short answers~\cite{zhang2024improve}. However, these machine-generated annotations are often too brief and \textcolor{improvered}{difficult} to segment semantically.

%% file: sec/3_method.tex
\begin{figure*}[t]
    \centering
    \includegraphics[width=\textwidth]{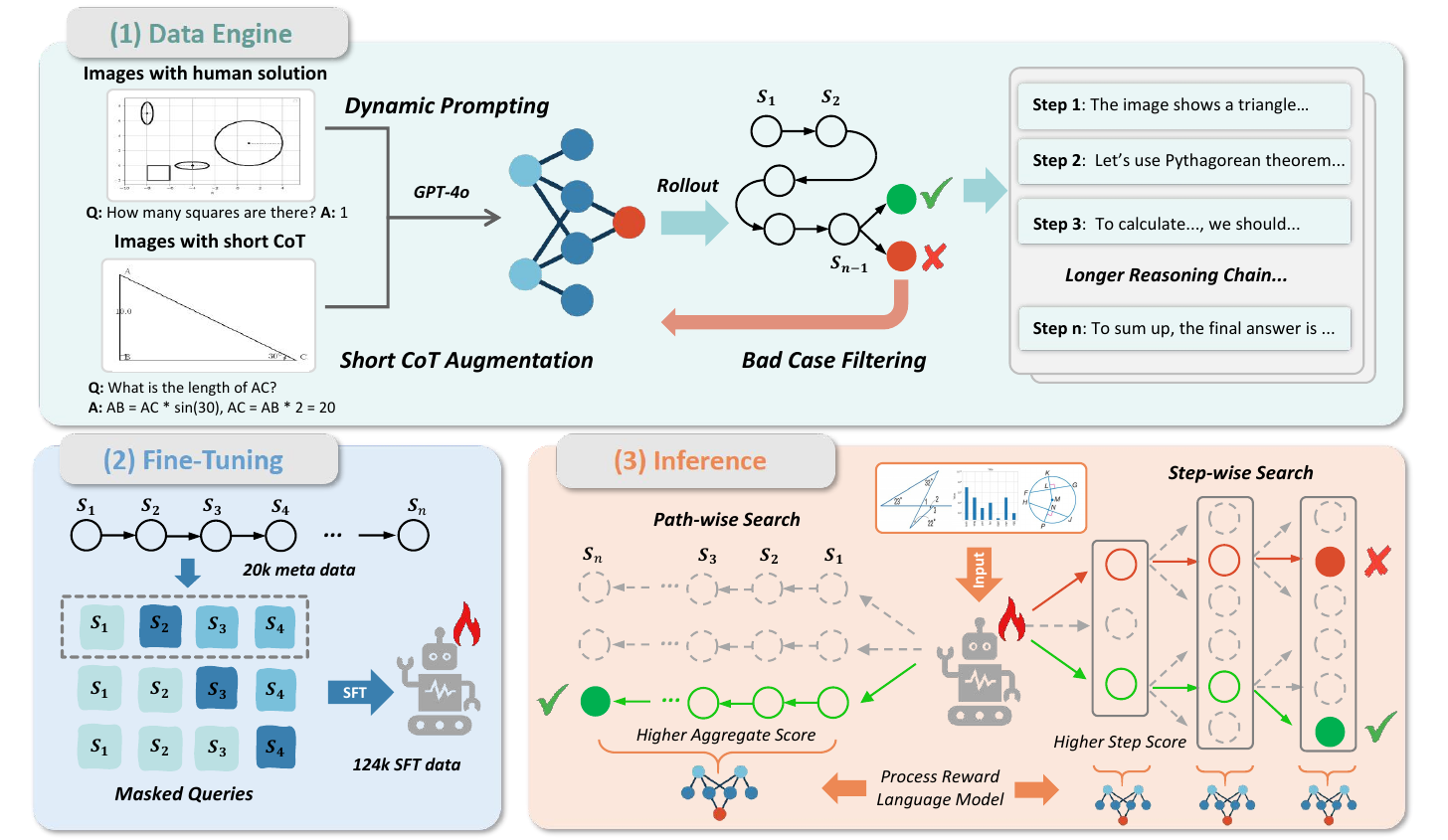} 
    \caption{The overview of AtomThink framework. We annotate and filter the open-source data with long CoT to generate atomic steps for fine-tuning and PRM training. During inference, step-wise or path-wise searching strategies can be applied to find optimal policies. Finally, the behavior distribution of GPT-4o is obtained through clustering with Kimi1.5, and an outcome-based method is employed for atomic step utilization evaluation.}
    \label{fig:framework}
\end{figure*}

\section{Method}
We present the details of AtomThink for promoting MLLM reasoning with self-structured CoT in this section. As shown in Figure \ref{fig:framework}, AtomThink consists of four key components\textcolor{improvered}{:} a self-structured reasoning mechanism (Sec.~\ref{seq:scot}), a data engine (Sec. \ref{sec:engine}), an atomic step fine-tuning process (Sec. \ref{sec:atomic_finetune})\textcolor{improvered}{, and} an atomic capability evaluation (Sec. \ref{sec:metric}). 

\subsection{Self-structured Chain-of-Thought}
\label{seq:scot}
To enable MLLMs to adaptively generate diverse reasoning paths in response to various problems \textcolor{improvered}{that mirror} human cognition, we propose an inference method based on Self-structured Chain-of-Thought (SCoT). In contrast to structured methodologies, our approach does not constrain the model to a fixed template of thought or a predefined sequence of reasoning steps \textcolor{improvered}{but instead} empowers the model to autonomously seek optimal reasoning behaviors at inference time.

\noindent\textbf{Multi-round Atomic Step Generation.} We commence by defining the minimal predictive action with semantic consistency as an \textit{Atomic Step}, which may constitute a single sentence or a combination thereof. Utilizing atomic steps as fundamental building blocks, we propose a multi-round prediction method to iteratively self-generate thought chains with dynamic structures. During the reasoning process, we prompt the model to predict only one minimal atomic step at a time \textcolor{improvered}{to focus} on the quality of each atomic step. Subsequently, the current prediction is appended to the historical reasoning steps and provided as contextual input for the next prediction cycle. Our reasoning template with SCoT is shown in Figure~\ref{fig:prompts_atomthink} \textcolor{improvered}{in} \textcolor{improvered}{Appendix.2}.

Due to the limitations of the model's instruction-following capability, we often observe hallucinations during reasoning \textcolor{improvered}{such as} repetitive atomic steps and duplicated sentences within steps \textcolor{improvered}{that can cause} the reasoning process to fall into self-referential loops and stagnation. Therefore, we use the following methods for anomaly detection and thought restart:

\begin{itemize}
  \item \textbf{Rule-based Filter:} We employ template matching and Jaccard similarity to \textcolor{improvered}{quantify} intra- and inter-step semantic repetition\textcolor{improvered}{and mitigate} looping phenomena. Given the observed potential for partial token mutations and imperfect matches in intra-step repetitions, we set the allowable repetition threshold at below 45\% while the inter-step repetition rate should remain under 98\%. Additionally, we define $max\_step\_length$ and $max\_length$ \textcolor{improvered}{parameters} to control the maximum length of a single atomic step and max response.
  \item \textbf{Temperature Accumulation:} Upon detection of an anomaly, we perform a single-step inference anew to replace the erroneous atomic step. To enhance \textcolor{improvered}{outcome diversity}, we incrementally increase the temperature with each error to simulate \textcolor{improvered}{the diverse thinking strategies} of human cognition.
\end{itemize}

\noindent\textbf{Policy Search with Process Reward Model. }
Given that the model spontaneously segments atomic steps during reasoning, a natural consideration is the introduction of a Process Reward Model (PRM) to further expand the search space for predictive actions. Unlike traditional token-based or sentence-based search strategies, we sample candidates using atomic steps as the fundamental unit. As there are many search strategies to generate candidate actions, we categorize the existing strategies into path-wise searching and step-wise searching. In path-wise search, we build upon prior work \cite{wang2024openr,snell2024scaling} by parallel sampling multiple paths and aggregating scores to find optimal solutions. We investigate the following two methods:
\begin{itemize}
  \item \textbf{Majority Voting: } It combines multiple reasoning paths by selecting the most frequent outcome across them \textcolor{improvered}{and assumes} that consensus across different paths is more likely to lead to \textcolor{improvered}{the correct answer}.
  \item \textbf{Best-of-N:} Given a generative MLLM, \textcolor{improvered}{the best-of-N sampling method} generates n candidate rollouts simultaneously and selects \textcolor{improvered}{the solution} with the highest score. The evaluation of candidate reasoning processes is determined by PRM, which employs three aggregation methods to map the dense scores to \textcolor{improvered}{the overall value} of \textcolor{improvered}{the entire path}: 1) The worst action: Compare the worst action among all candidate rollouts. It penalizes solutions with any weak action and is used to search \textcolor{improvered}{for reasoning that is} sensitive to errors. 2) The last action: The score is derived from \textcolor{improvered}{the prediction of the final answer during inference}. 3) Average score: It is calculated by averaging rewards of all the actions in a chain. The explainability and consistency of intermediate reasoning are emphasized here as \textcolor{improvered}{being as important as the outcome}.
\end{itemize}

Step-wise search strategies start with an initial path and incrementally expand the sampling space for each atomic action. Beam search and greedy strategies are applied to prune branches with low quality.
\begin{itemize}
  \item \textbf{Greedy Algorithm:} It focuses on making \textcolor{improvered}{the locally optimal choice} at each step of the reasoning process \textcolor{improvered}{by selecting} the best immediate action (step) based on \textcolor{improvered}{the current state} without considering future consequences. 
  \item \textbf{Beam Search:} It explores multiple branches at each action and maintains a fixed number of top candidates for each stage of reasoning \textcolor{improvered}{to balance} between exploring different paths and exploiting the most promising ones.
  \item \textbf{\textcolor{improvered}{Monte Carlo Tree Search (MCTS):}} \textcolor{improvered}{It explores the search space through four phases: selection, expansion, simulation, and backpropagation. It balances exploration and exploitation through Upper Confidence Bounds (UCB), enabling more informed decisions by learning from simulated outcomes rather than relying solely on immediate evaluations. We use a UCT with 1.414 and a maximum exploration steps with 1500.}
\end{itemize}

Table~\ref{tab:search} provides \textcolor{improvered}{a comparative experiment} of different policy search methods. In our main experiment, we \textcolor{improvered}{employ} a step-wise beam search to extend the inference time.

\begin{figure*}[th]
    \centering
    \includegraphics[width=\textwidth]{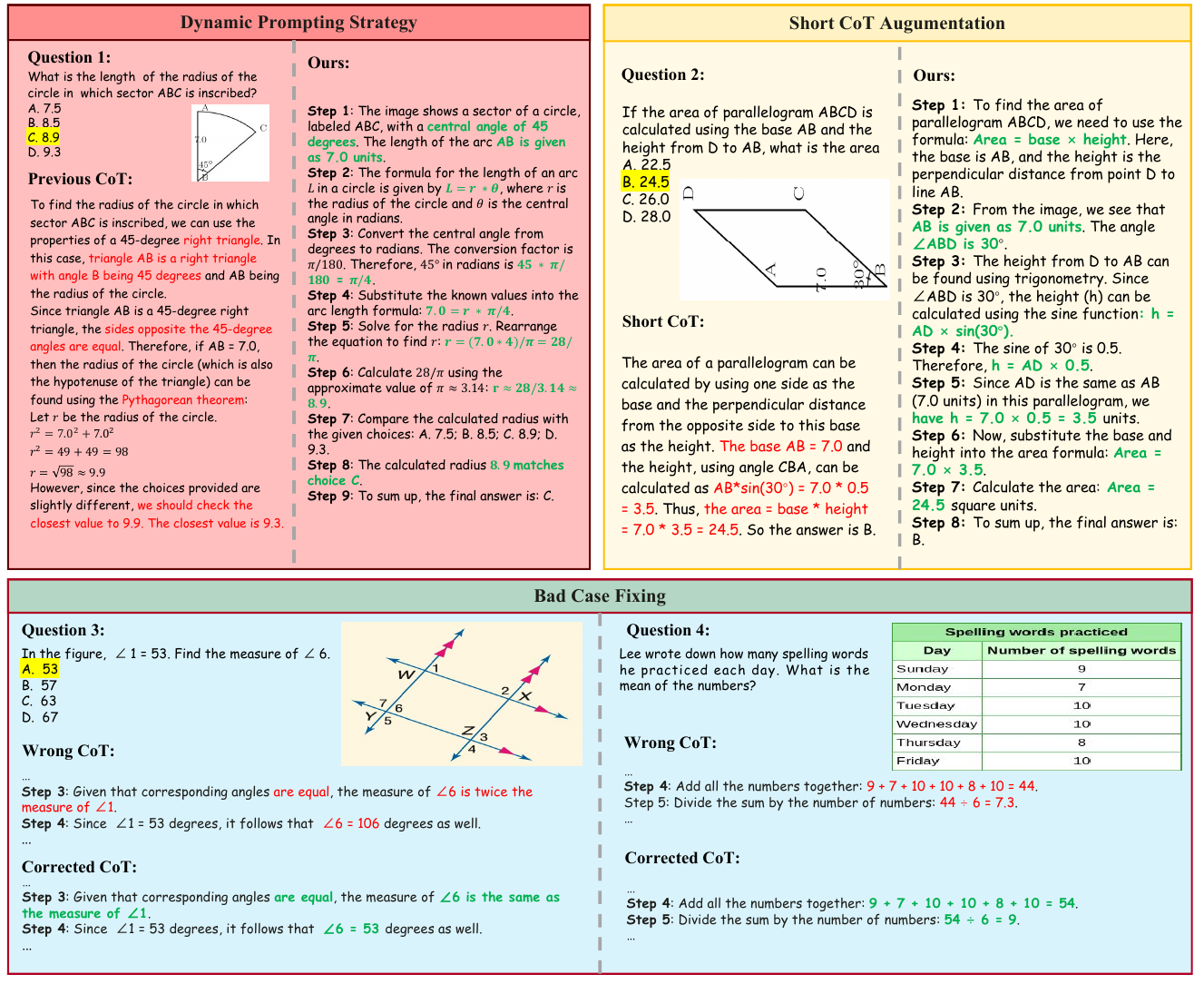} 
    \caption{Case study of our data engine to generate high quality CoT. Red and green characters denote incorrect and correct responses, respectively. Compared with vanilla CoT generated by GPT-4o, our dynamic prompting strategy exhibits fewer hallucinations in every atomic \textcolor{improvered}{step}. Utilizing existing short annotations, we can augment longer paths that encompass more details. Additionally, bad case filtering is applied to inspect low-quality noisy data within the automated pipeline.}
    \label{fig:data_engine}
\end{figure*}

\subsection{Data Engine}
\label{sec:engine}

Guiding MLLMs toward deep reasoning requires a substantial amount of high-quality CoT data. However, in the field of visual mathematics, the scarcity of publicly available datasets presents a considerable challenge. To overcome this, we develop an automated data engine capable of generating step-by-step long CoTs \textcolor{improvered}{that results in} our own atomic multimodal dataset \textcolor{improvered}{called} AMATH. Specifically, our data engine introduces a dynamic prompting strategy and \textcolor{improvered}{a short CoT augmentation strategy} to produce multi-step reasoning paths. Subsequently, we propose a difficulty scoring mechanism coupled with a secondary review strategy to sift through and filter out erroneous instances.


\noindent\textbf{Multimodal CoT Generation.} For long CoT generation, we propose two prompt-based methods:
\begin{itemize}
  \item \textbf{Dynamic Prompting.} Inspired by recent research \cite{g1}, we propose a dynamic prompt strategy for generating atomic inference steps. Specifically, our strategy drives a LLM to iteratively construct state-reasoning paths \textcolor{improvered}{where each path node} represents a reasoning step and encompasses the previous stage, the current state, and a possible action. The possible action includes continuing reasoning, verifying, and drawing \textcolor{improvered}{a conclusion}, which is determined by the LLM itself. The prompt is shown in \textcolor{improvered}{Appendix 2}. This strategy enhances reasoning performance \textcolor{improvered}{by enabling} \textcolor{improvered}{the model} to generate higher-quality CoT with reduced error rates, as demonstrated in Question 1 of Figure~\ref{fig:data_engine}.
  \item \textbf{Short CoT Augmentation.} To fully leverage existing short CoT annotations of VQA datasets, we also employ an MLLM to atomize and augment these annotations. This approach allows us to semantically segment an original reasoning process into multiple discrete steps \textcolor{improvered}{and focus} on solving a single atomic problem at each stage of the reasoning process. As shown in Question 2 of Figure~\ref{fig:data_engine}, we deconstruct the original reasoning pattern and incorporate detailed procedural descriptions.
\end{itemize}

\input{tables/dataset}
\input{tables/data_quality}

\noindent\textbf{Bad Case Filtering.} Due to the prevalence of substantial noise within the publicly available datasets, we first employ a difficulty scoring system to filter the questions \textcolor{improvered}{and subsequently use a} LLM  for a secondary review to eliminate erroneous CoTs.
\begin{itemize}
  \item \textbf{Difficulty Scoring.} To quantify the difficulty of questions, we employ Qwen2-VL-7B to sample $N$ candidates for each question \textcolor{improvered}{and use} the win rate of $N$ candidates as \textcolor{improvered}{the difficulty level} of the question ($N=10$ in our paper). To enhance the efficiency of training, we have removed most questions with a difficulty level of 0.
  \item \textbf{Secondary Review.} Upon the generation of CoT, we utilize GPT-4o to conduct \textcolor{improvered}{a secondary review} with a particular focus on the accuracy of atomic steps and the correctness of final answers. Furthermore, we engage two professional annotators to perform a sampling inspection of our dataset. As shown in Questions 3 and 4 of Figure~\ref{fig:data_engine}, this phase can correct or eliminate most flawed samples, such as calculation errors \textcolor{improvered}{and image recognition mistakes}.

\end{itemize}

\noindent\textbf{AMATH Dataset. } 
We sample multimodal reasoning data from CLEVR \cite{johnson2017clevr}, Geometry3K \cite{geo3k}, MAVIS \cite{zhang2024mavis}, TabMWP \cite{TabMWP}, GeomVerse \cite{kazemi2024geomverse}, Mathv360k \cite{shi2024math}, GeoQA+~\cite{chen2021geoqa} and IconQA~\cite{lu2021iconqa}. For GeomVerse and MAVIS, we conduct short CoT augmentation, while the rest are generated by dynamic prompts to produce multi-step reasoning. Table \ref{tab:dataset} illustrates the distribution of our data. In Table \ref{tab:diff_data_style}, we also evaluate the quality in a subset of 500 AMATH samples with GPT-4o scoring. \textcolor{improvered}{Additionally, we first use DeepSeek-V3.2 as an extractor to identify potentially problematic samples, followed by manual inspection of their reasoning process by three annotators (all PhD candidates with advanced mathematics education). Even though we observed rationale artifacts in some reasoning chains (depending on the instruction-following capability of the distilled model), the reasoning logic achieved an accuracy of up to 98.2\%. Compared with vanilla CoT, AMATH achieves the highest GPT-4o preference score, human accuracy and token length. We also identified some instances of step redundancy (11.9\%). The generation and filtration examples of our dataset are shown in Figure~\ref{fig:data_engine}.}

\subsection{Supervised Fine-Tuning}
\label{sec:atomic_finetune}

To fully exploit MLLMs for addressing \textcolor{improvered}{multimodal mathematical} problems, we conduct fine-tuning with atomic step-wise reasoning. We dissect CoTs from the metadata of AMATH into atomic steps and subsequently \textcolor{improvered}{employ} serialized masking to incrementally incorporate these into the historical reasoning steps \textcolor{improvered}{to generate} multiple training samples (denoted as AMATH-SFT) for supervised instruction fine-tuning.

\begin{figure}[t]
    \centering
    \includegraphics[width=0.48\textwidth]{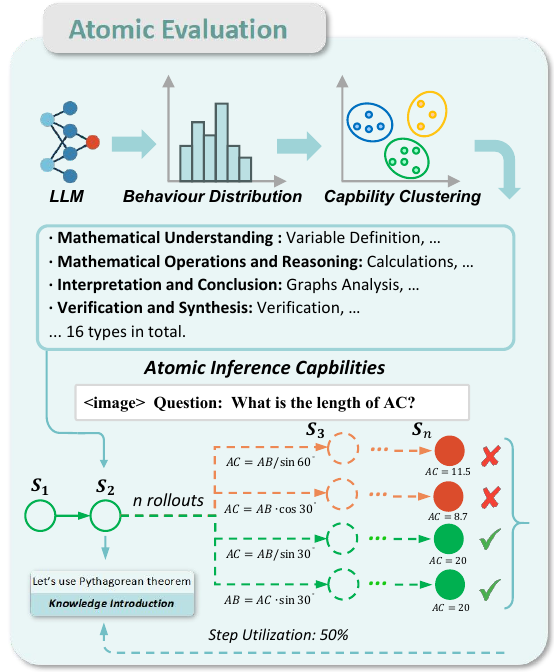} 
    \caption{Atomic capability evaluation. The capabilities are derived from the clustering of GPT-4o's behavior. By sampling each atomic step and evaluating the accessibility of the results, we assign a soft label that represents \textcolor{improvered}{the quality} of an atomic step.}
    \label{fig:evaluation}
\end{figure}

\subsection{Atomic Capability Evaluation}
\label{sec:metric}

Similar to human problem-solving processes, a SCoT may involve multiple reasoning abilities. However, traditional CoT methods do not focus on the ability to follow \textcolor{improvered}{individual reasoning steps} or provide fine-grained analyses of the underlying abilities. To address this gap, we develop an atomic capability evaluation strategy \textcolor{improvered}{that offers} a new analytical perspective for reasoning.

Our evaluation method aims to assess the mathematical capabilities of a target model from various perspectives, such as understanding, operations, and certifications. To this end, we first construct a canonical set of capabilities. As shown in Figure \ref{fig:pie_outline}, we collect the behavior distribution of GPT-4o on \textcolor{improvered}{the AMATH dataset} and use Kimi-1.5 to perform clustering \textcolor{improvered}{to yield} clusters \textcolor{improvered}{where each represents} a certain ability utilized by high-level intelligent models in solving mathematical problems. We consider each cluster as a set and let $Set(a)$ denote the cluster of an ability $a$.

We initially posit that models with superior atomic reasoning capabilities are more adept at leveraging recent contextual steps to further \textcolor{improvered}{derive} answers. Hence, we can quantify a certain reasoning ability of a model based on its average probability of reaching a correct answer with its rollouts sampled from the corresponding ability set. Specifically, assume a question has $n$ historical reasoning steps $S=\{s_i|i =1,...,n)\}$. We define the step utilization rate $u(S)$ as the probability of reaching an answer by continuing to reason based on $S$ \textcolor{improvered}{averaged over} $M$ sampled rollouts:
\begin{align}
u(S) = \frac{\sum^M_{m=1}\llbracket {r}_{m} \text{ is correct}\rrbracket}{M},
 \label{rate}
\end{align}
where $r_m$ is the $m$-th rollout \textcolor{improvered}{and $\llbracket P \rrbracket$ denotes the Iverson bracket which equals to 1 if predicate $P$ holds and otherwise 0.} Subsequently, we calculate the utilization rates of different historical steps and map the corresponding $S$ back to the set of atomic capabilities. We compute the average utilization rate for each category in the ability set to represent the model's atomic reasoning capability, which can be represented as follows\textcolor{improvered}{:}
 \begin{align}
 Score(a) = \frac{1}{{|Set(a)|}}\sum\limits_{{S_k} \in Set(a)} {u({S_k})} .
 \label{ability_score}
\end{align}
In our experiments, we \textcolor{improvered}{select} 160 samples from an out-of-distribution mathematical dataset (R1V-Stratos \cite{yu25r1vision}) to construct a test set for atomic capability evaluation.


\begin{figure*}[t]
    \caption{Running case of CoT and SCoT. SCoT inference pipeline automatically concatenates historical steps to indicate the model's next step of thinking. \textcolor{improvered}{The model} can focus more on the previous steps, which also allows us to perform step-wise search methods.}
    \centering
    \includegraphics[width=\textwidth]{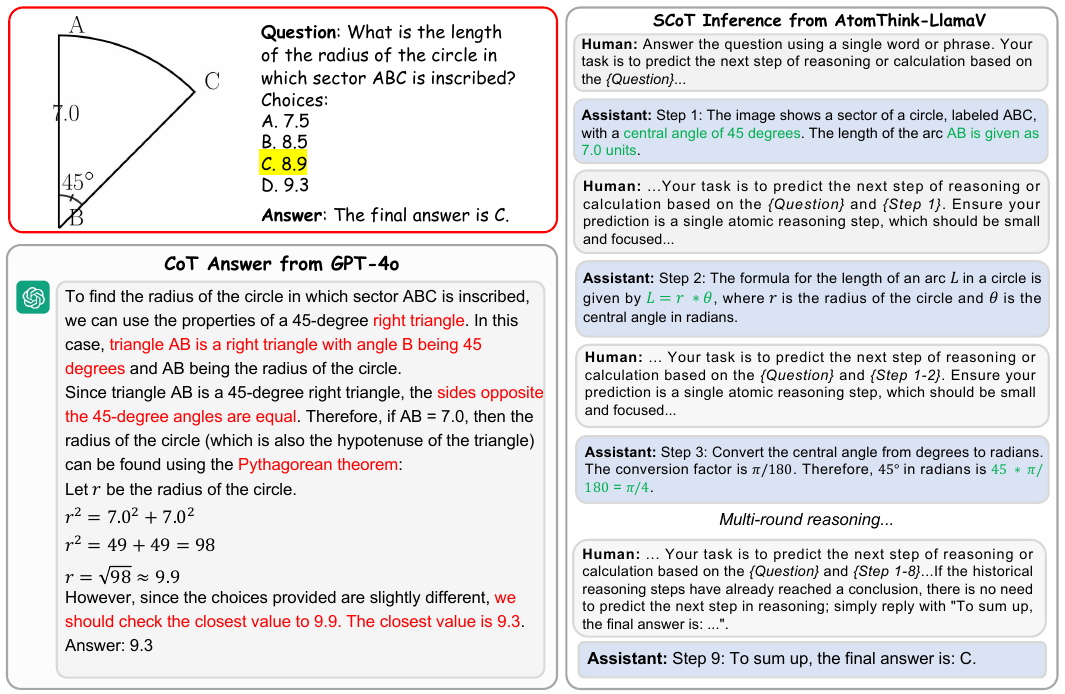} 
    \label{fig:running_case}
\end{figure*}

%% file: tables/dataset.tex
\begin{table}[ht]
\caption{Data composition of our AMATH. 20K VQA samples are applied to generate 124K SFT data with intermediate atomic steps.}
\centering
\resizebox{\columnwidth}{!}{  
\begin{tabular}{l c c c}
\toprule
\textbf{Source} & \textbf{AMATH-Metadata} & \textbf{AMATH-SFT} \\
\midrule
CLEVR        & 2056  & 11.9K   \\
Geometry3K   & 1224  & 9.3K  \\
MAVIS        & 1685  & 11.4K  \\
TabMWP       & 2643  & 16.3K  \\
GeomVerse    & 1347  & 9.9K \\
MathV360K    & 5632 & 31.6K  \\
GeoQA+       & 2222  & 15.5K  \\
IconQA       & 3199  & 18.1K \\
\midrule
Total        & 20008 & 124K \\
\bottomrule
\end{tabular}
}
\label{tab:dataset}
\end{table}

%% file: tables/data_quality.tex
\begin{table}[ht]
\centering
\caption{Comparison of different datasets. \textcolor{improvered}{We randomly sample 500 CoT examples for analysis.
Avg. Length: Average length of CoTs;
GPT Score: Use GPT-4o to score the quality of CoTs.
Redundancy: Reasoning redundancy verified by annotators.}
}
\label{tab:diff_data_style}
\resizebox{\columnwidth}{!}{ 
\begin{tabular}{c|c|c|c|c}
\hline
\textbf{Data} & \textbf{Avg. Length} & \textbf{GPT Score} & \textcolor{improvered}{\textbf{Accuracy}} & \textcolor{improvered}{\textbf{Redundancy}} \\
\hline
PRM800k & 1245.4 & 84.1 & - & - \\
Direct & 3.6 & 1.5 & 100 & - \\
Vanilla CoT & 670.5 & 79.6 & 92.8 &  \\
AMATH(Ours) & 849.8 & 89.4 & 98.2 & 11.9\\
\hline
\end{tabular}}
\end{table}

%% file: sec/4_exp.tex
\section{Experiment}
\label{sec:exp}

\input{tables/main_exp}

\input{tables/general_exp}

\subsection{Setup}
\noindent\textbf{Baselines.} Our main experiments utilize two different open-source MLLMs, including LLaVA1.5-7B~\cite{liu2024visual} and Llama3.2-11B-Vision~\cite{llama32v}. 
LLaVA1.5-7B connects the CLIP-ViT-L-336px visual encoder with Vicuna1.5-7B~\cite{zheng2023judging} language model via multilayer perceptron (MLP) projection layer, excelling in academically oriented visual question answering (VQA) tasks. Llama3.2-11B-Vision extends Llama 3.1 language model~\cite{dubey2024llama} by integrating a visual adapter through cross-attention layers, demonstrating strong multimodal reasoning capabilities.
With a subset of 100K multimodal question-answer pairs sampled from LLaVA-665K~\cite{liu2024visual}, we post-train full parameters of their language models, projectors and vision encoder as baselines.  We use a learning rate of 2e-6 and a batch size of 128 to fine-tune them for one epoch. The maximize context length is set to 4096 tokens. Specifically, we utilize the Llama-factory~\cite{zheng2024llamafactory} framework to train the models. In SFT stage, the AMATH-SFT dataset proposed in Section \ref{sec:engine}, is incorporated to introduce atomic reasoning capabilities. In addition, we further fine-tune LLaVA-Llama3-8B and EMOVA-8B models using AMATH-SFT for supplementary experiments. Detailed training parameters are provided in \textcolor{improvered}{Appendix.1}.
We select 10 popular MLLMs for comparison, including Claude 3.5 Sonnet~\cite{claude}, OpenAI’s o1~\cite{openai2024_o1}, 4o~\cite{openai2024_4o}, 4v~\cite{openai2024_4v}, as well as LLava-NeXT-34B~\cite{liu2024llavanext}, InternLM-XComposer2~\cite{zhang2024mavis}, Qwen-VL-Plus~\cite{bai2023qwen}, LLaVA-1.5-13B~\cite{liu2024visual}, LlamaV-o1-11B~\cite{llamavo1} and LLaVA-CoT-11B~\cite{llavacot}.
 
\noindent\textbf{Evaluation Protocol.} To assess the effectiveness of our method in enhancing multimodal reasoning capabilities, we conduct experiments across  \textcolor{improvered}{different tasks, including 4 mathematical benchmarks (MathVista~\cite{lu2023mathvista}, MathVerse~\cite{zhang2025mathverse}, MathVision~\cite{wang2024measuring}, MathVision~\cite{wang2024measuring} and WeMath~\cite{WeMath}), 4 scientific benchmarks (ScienceQA~\cite{ScienceQA}, AI2D~\cite{AI2D}, MMMU~\cite{MMMU} and HLE~\cite{phan2025humanity}) and 3 general benchmarks (DocVQA~\cite{DocVQA}, ChartQA~\cite{ChartQA} and TextVQA~\cite{TextVQA}).}  MathVista, a publicly available benchmark encompassing both general-targeted and mathematics-targeted domains. Additionally, MathVerse is introduced to assess model's sensitivity to mathematical graphs. MathVision, a benchmark encompassing a diverse range of mathematical problem complexities, is also incorporated into experiments to specifically evaluate the dynamic variations in our atomic steps. \textcolor{improvered}{WeMath investigates the model's answering mechanism by decomposing complex multi-concept problems into atomic sub-problems. For general benchmarks, DocVQA evaluates document understanding through visual question answering on scanned documents, requiring models to extract and reason over textual information from various document layouts. ChartQA assesses the ability to understand and reason about data visualizations such as bar charts, line graphs, and pie charts. TextVQA challenges models to answer questions that require reading and reasoning about text present in natural images. ScienceQA is a multimodal science question answering dataset covering natural science, social science, and language science topics with diverse question types. AI2D is a diagram understanding benchmark focusing on K-12 science diagrams.} MMMU \textcolor{improvered}{further improve the difficulty to college-level. As a cross-disciplinary task, it features 11.5K diverse questions spanning 6 core disciplines and 30+ image types.}  We also introduce Humanity's Last Exam (HLE), one of the most challenging benchmark, to assess model's reasoning capabilities under extremely difficult conditions. \textcolor{improvered}{Evaluation on MathVista, MathVerse, MathVision and ScienceQA is conducted using GPT-4o~\cite{openai2024_4o} as a judge, while the remaining benchmarks are evaluated through template matching.}

Our evaluations include four inference settings, including \textbf{Direct}, \textbf{CoT}, \textbf{SCoT}, and \textbf{SCoT w/ PRM}. In \textbf{Direct} setting, we prompt the model to generate a concise final answer. In \textbf{CoT}, models are instructed to answer the question through step-by-step reasoning. For the Direct and CoT evaluations, we use prompts from lmms-eval~\cite{zhang2024lmmsevalrealitycheckevaluation,lmms_eval2024}. Our AtomThink-models support two additional settings: \textbf{SCoT} and \textbf{SCoT w/ PRM}. In SCoT, our models follow a single, atomic reasoning path based purely on their learned policies, without employing any supplementary search strategies. 
In SCoT w/PRM, we directly utilize Qwen2.5-Math-PRM-7B~\cite{prmlessons} to provide high-quality step-wise rewards for LLaVA1.5-7B and Llama3.2-Vision-11B. Additionally, we fine-tune a PRM based on Math-psa-7B~\cite{wang2024openr} to offer process reward. This model employs the AMATH-Metadata and a 20k-sized subset of PRM800K~\cite{lightman2023let} as seed data, which is used to supervise the LLaVA-Llama-8B and EMOVA-8B models. In step-wise beam search, a window of 3 and candidate number of 2 are utilized. In other policy search policies we use candidate number of 3. During the search process, the temperature for each step is initialized at 0 and incremented by 0.5 with each candidate sampling to enhance diversity.

    

\begin{figure*}[t]
    \centering
    \label{fig:combined}
    
    \begin{minipage}{0.48\textwidth}
        \centering
        \includegraphics[width=\linewidth]{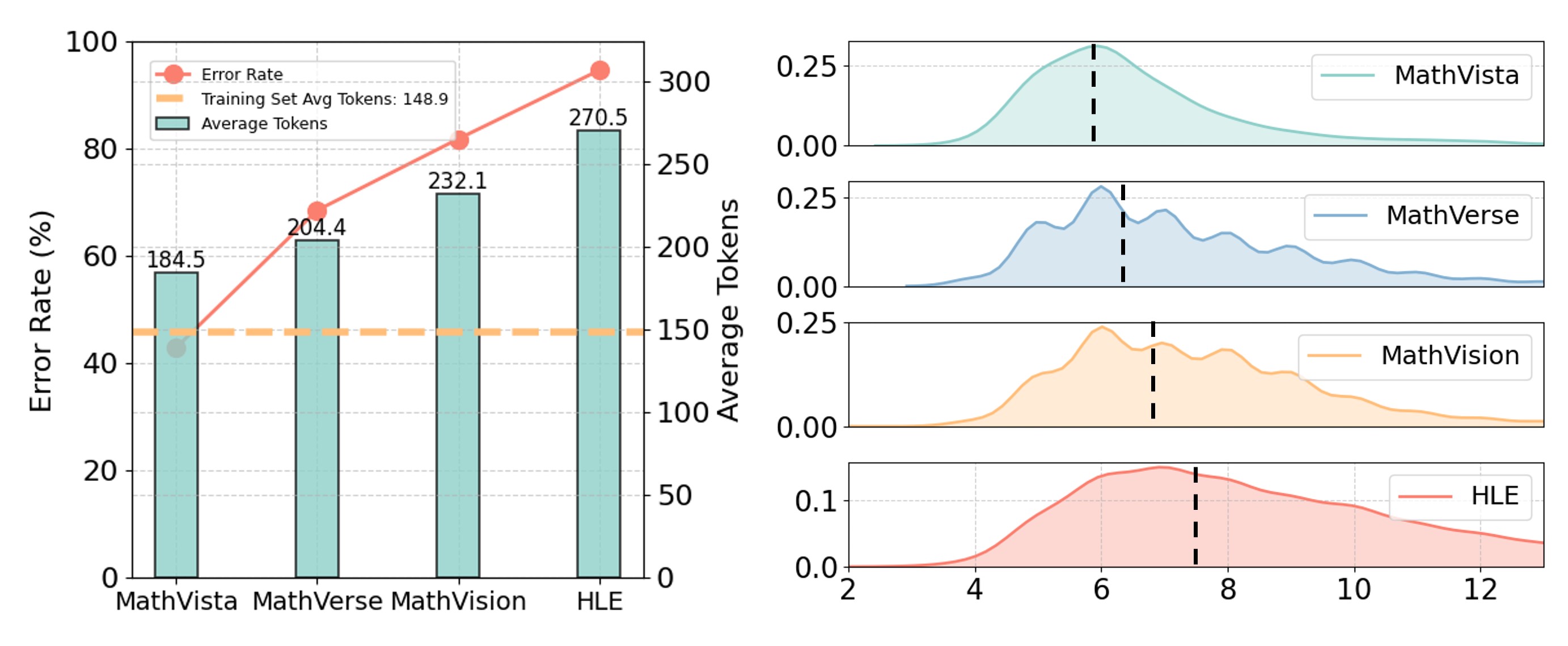} 
        \caption{Comparison of the average response length in AtomThink-LlamaV over benchmarks with different complexity. (a) As tasks become more challenging, the model proactively utilizes more tokens. (b) The proportion of longer CoT containing a greater number of atomic steps increases in outputs.}
        \label{fig:step_length}
    \end{minipage}
    \hfill 
    \begin{minipage}{0.48\textwidth}
        \centering
        \includegraphics[width=\linewidth]{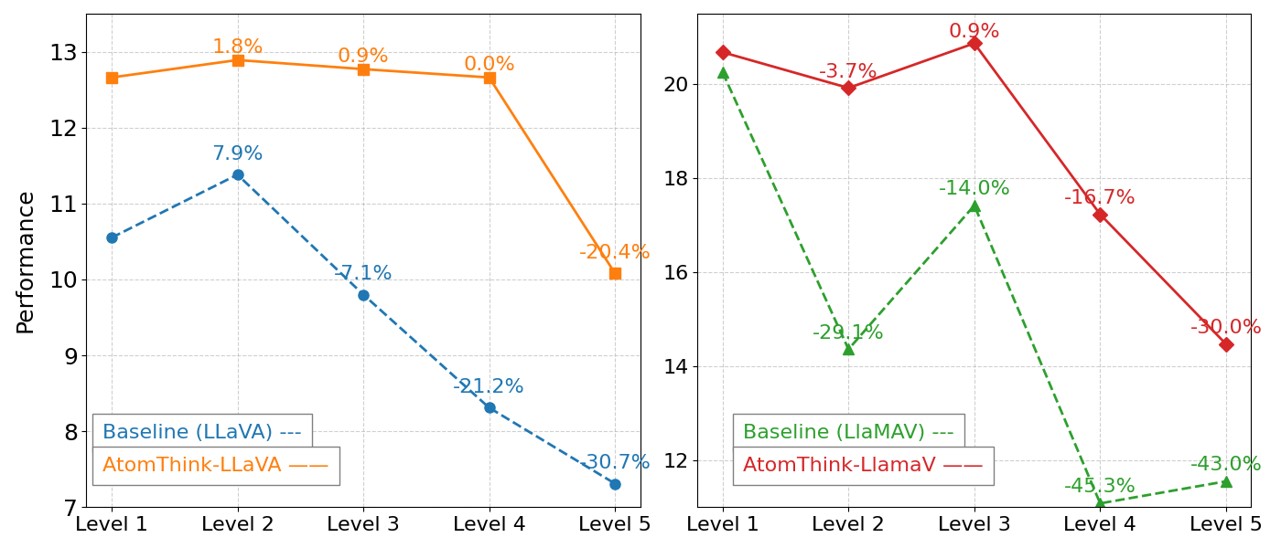} 
        \caption{MathVision-mini accuracy in diverse difficulty level subsets. A higher level signifies increased difficulty. The performance decline margin of AtomThink modes are more narrow (-20.4\% v.s. -30.7\% in LLaVA1.5, -30\% v.s. -43.0\% in LlamaV).}
        \label{fig:difficulty_performance}
    \end{minipage}
\end{figure*}

\subsection{Main Results}

\noindent\textbf{\textcolor{improvered}{Mathematical Reasoning.}} Figure~\ref{fig:running_case} shows a running case of CoT and SCoT. In Table~\ref{tab:main}, our AtomThink framework is applied to train LLaVA1.5-7B and Llama3.2-Vision-11B, yielding consistent performance improvements over the original models. With Self-structured CoT, the accuracy of AtomThink-LLaVA can be enhanced by 4.4\% and 3.4\% in MathVerse and MathVision, respectively. In a larger vision understanding model, AtomThink-LlamaV gains a higher improvement by 9.6\% and 8.2\%. When combined with step-wise beam search and process reward model, AtomThink-LlamaV achieves a new state-of-the-art on MathVista, surpassing GPT-4V and narrowing the gap between MLLMs and human performance.  \textcolor{improvered}{On WeMath benchmark, which places greater emphasis on sub-problem solving, AtomThink series achieve remarkable performance improvements (up to 38\%), attribute to our incorporation of foundational knowledge and focus on single-step reasoning.}


\noindent\textbf{\textcolor{improvered}{Cross-domain Generalization.}} \textcolor{improvered}{Although our constructed AMATH dataset does not contain knowledge from physics, chemistry, and other scientific domains, the reasoning capabilities still generalize to scientific and general tasks. As shown in Table~\ref{tab:main2}, AtomThink-LlamaV achieves gains across multiple domains: 25.3\% on ChartQA, 20.2\% on ScienceQA, 12.2\% on TextVQA, and 6.4\% on DocVQA. Notably, the improvements on general reasoning tasks (particularly ChartQA and TextVQA) surpass those observed on in-domain mathematical benchmarks. This improvement in generalization capability stems from the effectiveness of our high-quality data fine-tuning. Furthermore, SCoT forces model to revisit visual tokens multiple times by inheriting historical reasoning paths, which may reduce the recognition bias that could exist in a single input pass.}

\noindent\textbf{\textcolor{improvered}{Further Improvement with PRM.}} \textcolor{improvered}{Moreover, our investigation reveals that employing external PRM for reasoning path search yields varying gains across models and tasks. For instance, AtomThink-LLaVA shows a 2.9\% improvement on MathVista with PRM, while both models exhibit modest degradation on HLE (-1.3\% and -0.9\%). This may stem from PRM providing excessive yet ineffective supervision for challenging problems, coupled with its sensitivity to varying MLLM output styles.
Additionally, integrating PRM during inference amplifies cross-domain benefits. On ChartQA, the performance gain increases from 18.7\% (SCoT alone) to 25.3\% (SCoT w/ PRM), suggesting that step-wise verification enhances reasoning quality on simpler tasks. However, this generalization is not uniform. Scientific benchmarks like ScienceQA and HLE show performance degradation (-0.1\% and -0.9\%), likely due to their reliance on domain-specific knowledge absent from PRM's training data distribution.}

\textcolor{improvered}{In summary, the AtomThink framework demonstrates significant performance improvements across multimodal mathematical and cross-domain tasks through its atomic step-based reasoning approach and Self-structured Chain of Thought generation. The architecture exhibits natural compatibility with process reward models, enabling further performance gains.}

\begin{figure}[t]
    \centering
    \includegraphics[width=0.5\textwidth]{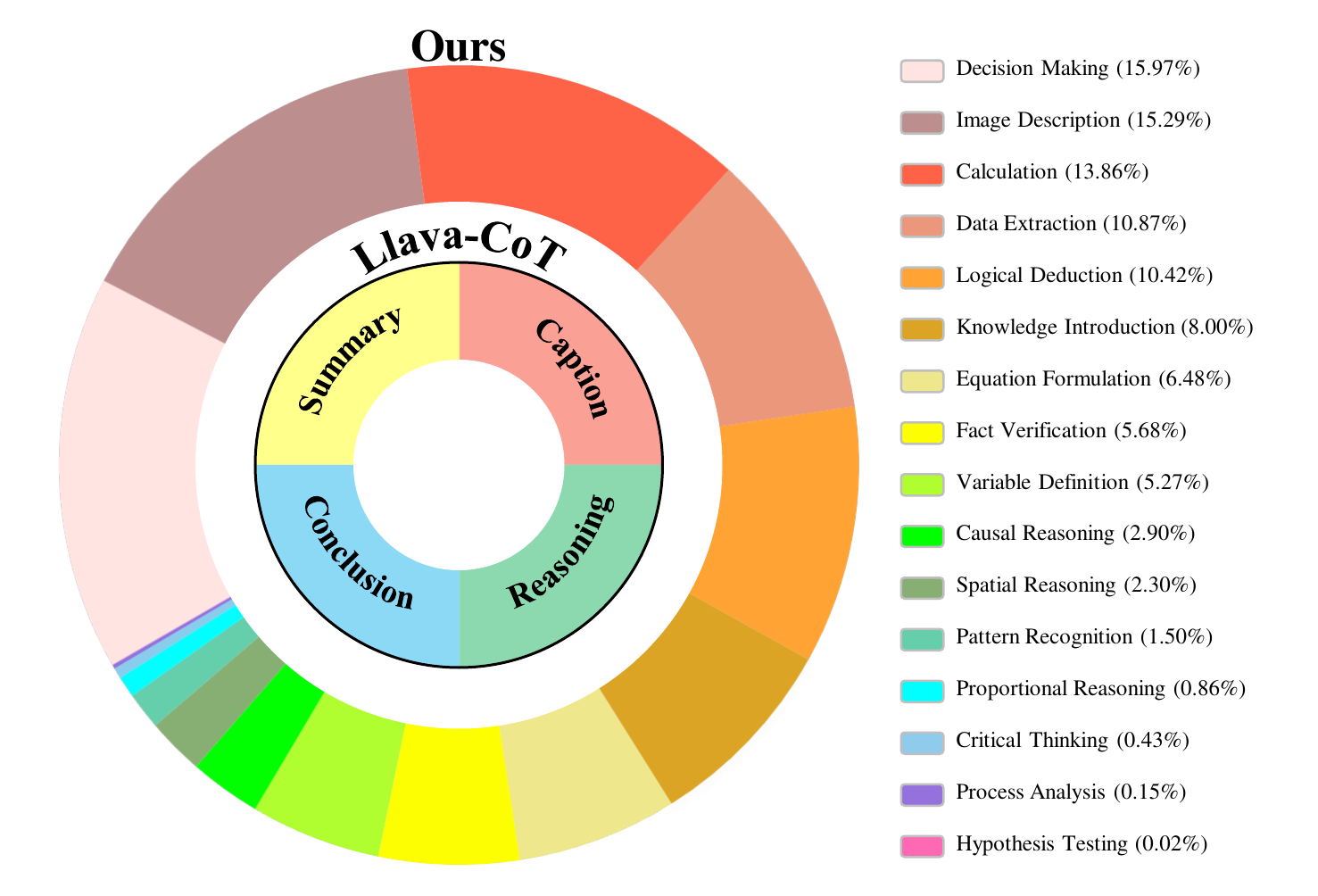}
    \caption{Reasoning step distribution of AtomThink-LlamaV and LLaVA-CoT.}
    \label{fig:pie_outline}
\end{figure}

\begin{figure}[t]
    \centering
        \centering
        \includegraphics[width=0.5\textwidth]{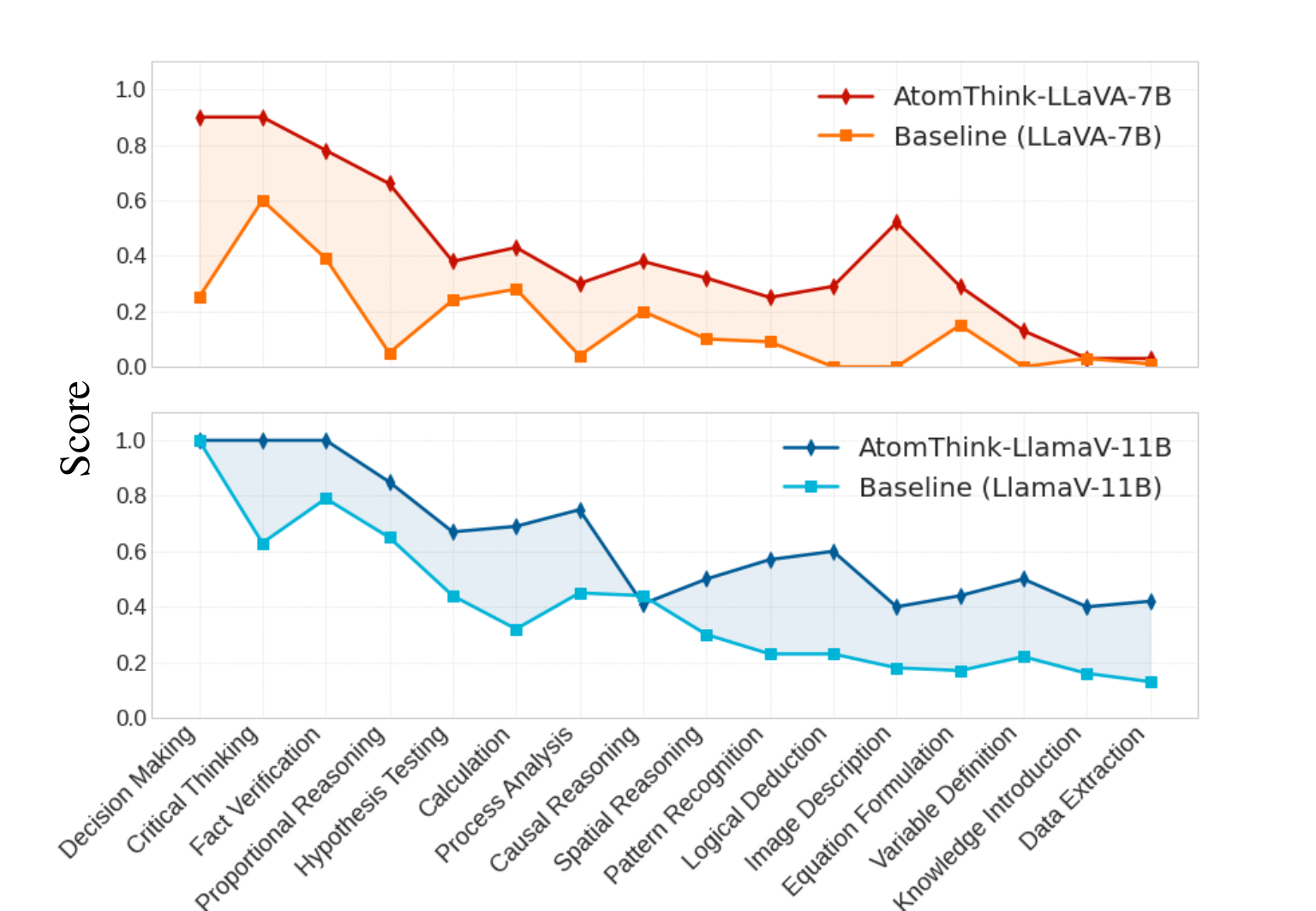}
        \caption{Utilization efficiency evaluation across atomic capabilities.}
        \label{fig:score_outline}
\end{figure}

\subsection{Scaling Reasoning According to Difficulty}
To assess the variation in the length of SCoT under differential difficulty, we present the output distribution of AtomThink-LlamaV across four benchmarks in Figure~\ref{fig:step_length}. The ascending error rates indicate a sequential increase in benchmark difficulty. In subplot (a), despite the train set distribution averages 148.9 tokens per reasoning step, the model demonstrates difficulty-adaptive behavior by producing longer reasoning chains for more challenging test problems (from 184.5 to 270.5). This suggests that the model is not merely fitting the training data but is instead exhibiting an emergent ability to autonomously explore the depth of reasoning. Subplot (b) illustrates that the model predominantly generates SCoT sequences containing 4-8 steps. More challenging benchmarks (e.g., HLE) exhibit a notable presence of extended reasoning chains exceeding 12 steps. The results reflect that even without human intervention, the model employs a greater number of atomic steps to address more complex problems.

Beyond evaluating models' adaptive capabilities across benchmarks of varying difficulty, we further leverage MathVision's pre-defined difficulty labels to conduct a systematic comparison of performance differentials. In Figure~\ref{fig:difficulty_performance}, both models demonstrate robust performance on low-to-medium difficulty problems when using our method. Specifically, AtomThink-LLaVA maintains stable accuracy (less then 0.5\%) across Level 1 to Level 4, while AtomThink-LlamaV achieves 0.9\% improvement from Level 1 to Level 3 problems. Collectively, the integration of AtomThink enhances models' adaptability to reasoning questions of varying complexity levels, as evidenced by reduced accuracy decline margin across difficulty spectrums.

\input{tables/llavacot_and_data_scaling}
\input{tables/test_time_scaling}

\subsection{Autonomous Generation of Diverse Structures}
 We cluster the reasoning behaviors of GPT-4o into 16 categories and collect the distribution of atomic steps produced by AtomThink on the Stratos160 test set. The results in Figure~\ref{fig:pie_outline} demonstrate that, compared to structured output (LLaVA-CoT), our SCoT exhibits a more diverse range of reasoning structures. Among all categories, there are some predominant reasoning patterns, e.g. Decision Making (15.97\%), Image Description (15.29\%), Calculation (13.86\%) and Data Extraction (10.87\%). These high-frequency operations constitute the core cognitive processes underlying the model's problem-solving methodology. With the enhanced visual understanding abilities, the model also displays specific behaviors such as Causal Reasoning (2.9\%) and Spatial Reasoning (2.3\%). Intriguingly, some outputs spontaneously exhibit self-verification behaviors in reasoning process, as exemplified by 5.68\% of Fact Verification and 0.02\% of Hypothesis Testing.

\subsection{Data Utilization and Reasoning Efficiency}
Table~\ref{tab:llavacot} provides a comprehensive performance comparison between our method and recently proposed Structured CoT approaches (LLaVA-CoT~\cite{llavacot} and LlamaV-o1~\cite{llamavo1}). AtomThink demonstrates across-the-board improvements in accuracy, data utilization efficiency, output token efficiency and inference latency. By utilizing only one-fifth of VQA samples, we achieve a 3.6\% improvement on MathVista. Furthermore, due to our ability to provide concise responses to simpler questions, we reduce output tokens by 87.8\% and inference time by 85.3\% per sample. Even with test-time scaling using PRM, our approach achieves significant reductions of 44.4\% in token count and 33.4\% in inference latency, demonstrating the method's strong potential for resource-constrained deployment scenarios.

\subsection{Scaling Law in Data and Test-time}
Previous research has found that scaling up data and test-time computations can enhance reasoning in language models. Our result also discovers that this scaling law persists in multimodal models. \textcolor{improvered}{Table~\ref{tab:data_scaling} presents the experimental results for different scales and ratios of the base dataset (LLaVA-Instruct, from 0k to 400k) and additional dataset (AMATH-SFT, from 0k to 124k). As the AMATH data increases, model's accuracy on the MathVision-mini benchmark steadily improves from 9.67\% to 12.45\%. Performance degradation occurs when fine-tuning without incorporating the base dataset, which we attribute to the substantial distribution shift between AMATH-SFT and the original training data.} As detailed in Table~\ref{tab:tt_scaling}, we employ a step-wise Best-of-N strategy, linearly increasing reasoning time by increasing the number of candidates for each reasoning step. With no search strategy (Candidate=0), baseline accuracy is 13.9\%. As the number of candidates increased to 3, accuracy significantly rises to 23.3\%, with each additional candidate contributing an average improvement of 3.1\%. Concurrently, the average output token count increases from 2.3 to 822.3, reflecting that model engages in "slower" and more in-depth thinking to explore better solution paths.

\input{tables/more_models}

\begin{figure}[th]
    \centering
    \includegraphics[width=0.5\textwidth]{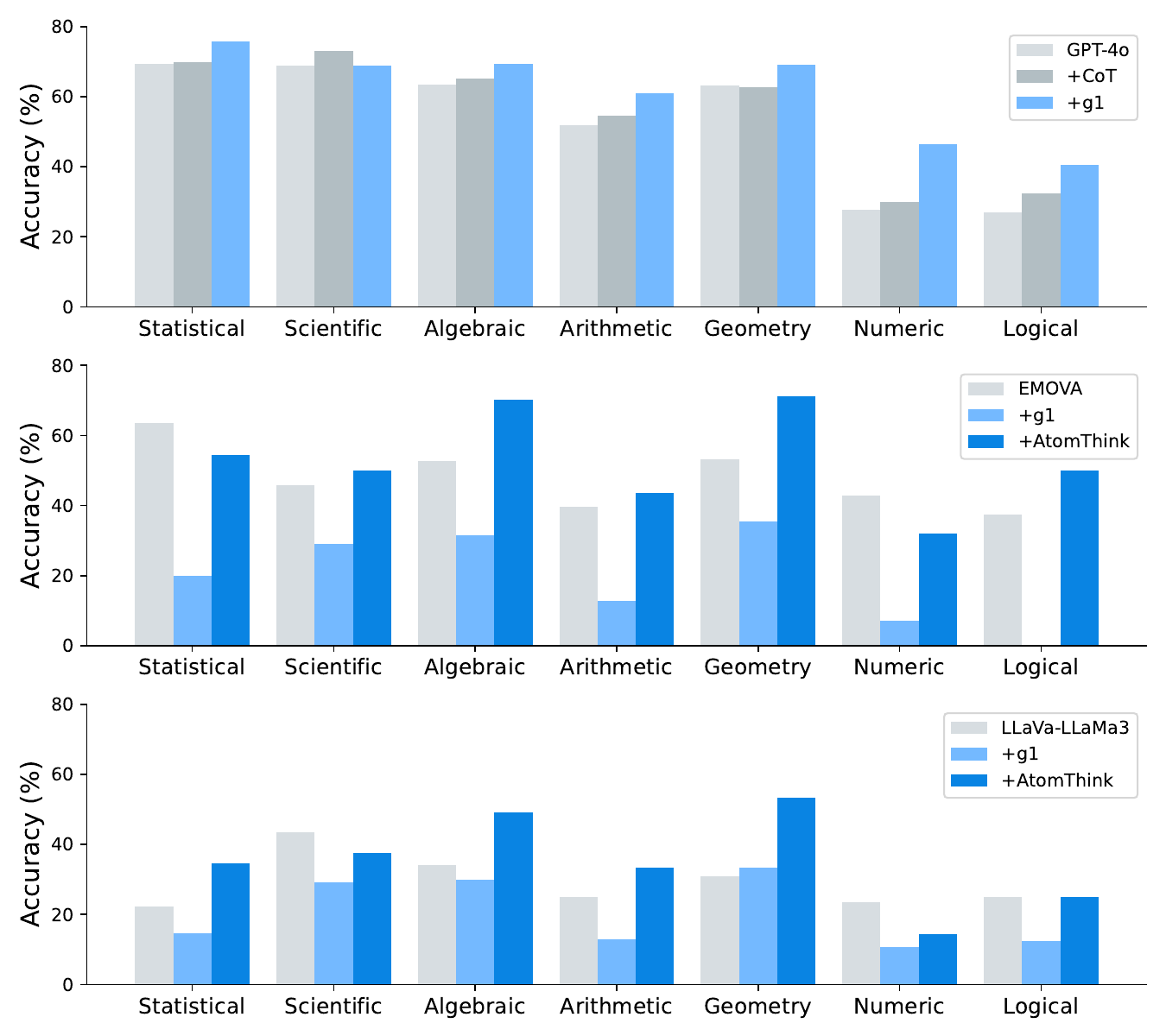} 
    \caption{Comparison to CoT and g1 in MathVista subsets. In contrast to the declining trend observed in g1, AtomThink outperforms the baseline across most subsets.}
    \label{fig:g1}
\end{figure}

\subsection{Cross-model Applicability}
To validate the versatility and cross-model applicability of our framework, we apply it to more open-source models with varying architectures. As shown in Table~\ref{tab:app_main}, AtomThink consistently delivered significant performance gains across all base models. For instance, AtomThink-EMOVA improves accuracy on MathVerse from 25.7\% (with standard CoT) to 40.5\% (with SCoT and PRM). In addition, it exhibits a certain degree of performance decline (3.5\%) in general tasks requiring common sense capabilities, which may be attributed to the model's originally weaker CoT reasoning ability. In Figure~\ref{fig:g1}, we compare AtomThink with the state-of-the-art open-source inference strategy, g1\footnote{https://github.com/bklieger-groq/g1}, which employs dynamic prompting to make model focus on single step reflection. In GPT-4o, direct application of g1 for multi-turn reasoning yields a greater improvement over Chain-of-Thought, particularly in numeric and geometric tasks. However, due to the reliance on the inherent reasoning capabilities of large-scale language models, its performance significantly degrades on smaller models such as EMOVA-8B and LLaVA-Llama3-8B. In contrast, our AtomThink framework consistently enhances the performance of these MLLMs.

\subsection{Effects on Policy Search Strategies}
In Table~\ref{tab:search}, we evaluate the impact of direct output, path-wise search and step-wise search strategies on MatVista and MathVerse using a subset of 300 samples from each. Results show that even without additional computation, AtomThink-EMOVA’s direct prediction accuracy outperforms the original, with improvements of 1.3\%, 1.52\%, and 2.4\%, respectively. The path-wise search method, BoN-Avg, achieves the highest accuracy of 58.68\% on the MathVista mathematical tasks, although it experienced a drop on general problems. \textcolor{improvered}{For step-wise methods, MCTS shows slight improvement over beam search on math but exhibits modest degradation on MathVista-General, indicating diminishing marginal returns in reasoning performance.} Meanwhile, both greedy algorithm and beam search show balanced performance across all benchmarks, with the generalization gap between math and general tasks being notably smaller than that of path-wise search. These results indicate that different search strategies can significantly influence reasoning outcomes. Due to AtomThink's enhancement of the diversity in model's reasoning actions, models can achieves consistent improvements across various search strategies.

\input{tables/search_method}

\input{tables/ablation}

\input{tables/prm}

\subsection{Ablation Study}
To systematically analyze the contributions of each core component of AtomThink framework, we conduct a series of ablation studies using MathVista in Table~\ref{tab:ablation}. We evaluate the impact of AMATH-SFT dataset, Self-structured Chain-of-Thought (SCoT) inference paradigm, and the Process Reward Model (PRM)-guided search strategy. In Llama3.2-Vision-11B, accuracy is improved from 44.3\% to 57.1\% with AMATH-SFT training set. This significant gap demonstrates that our atomic-step fine-tuning is crucial for model to learn how to generate effective reasoning behaviours. Different inference strategies also have impact on performance. Accuracy of Llama3.2-Vision-11B using direct output and CoT is only 47.5\% and 50.4\%, whereas SCoT improves it to 57.1\%. With PRM, the accuracy on AtomThink-LlamaV improves further from 57.1\% to 58.4\%. The improvement is even more pronounced on the AtomThink-LLaVA model, increasing from 29.4\% to 32.1\%. This indicates that by performing fine-grained quality assessment and selection for each step, PRM can effectively correct potential reasoning biases

  \begin{figure}[th]
    \centering
    \includegraphics[width=0.5\textwidth]{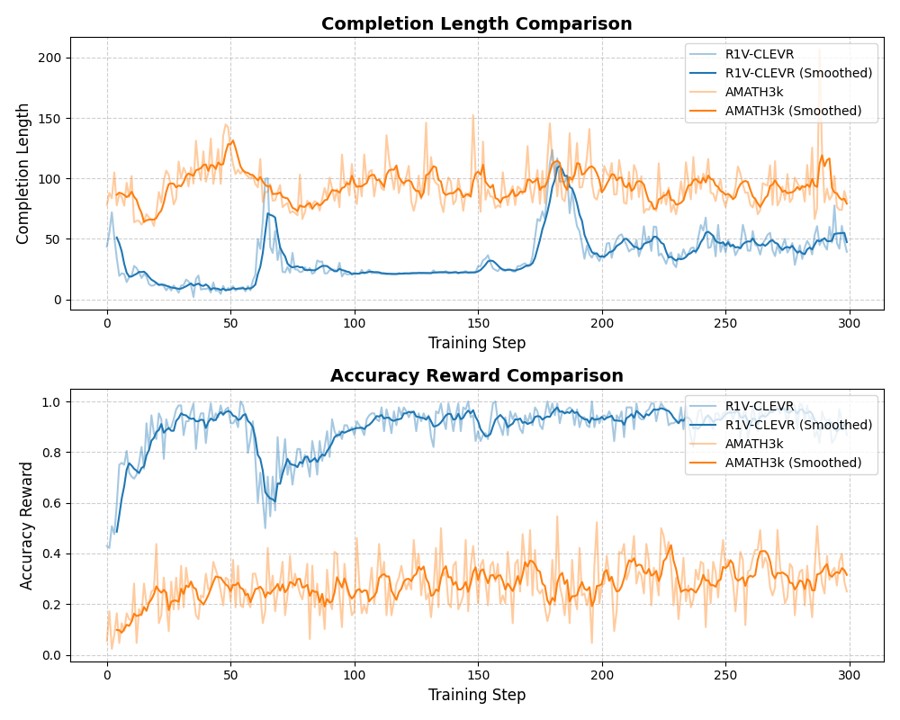} 
    \caption{Comparison with a DeepSeek-R1 like framework using reinforcement learning. A 3k subset of AMATH is sampled for fair comparison.}
    \label{fig:grpo}
\end{figure}

\begin{figure*}[th]
    \centering
    \includegraphics[width=0.95\textwidth]{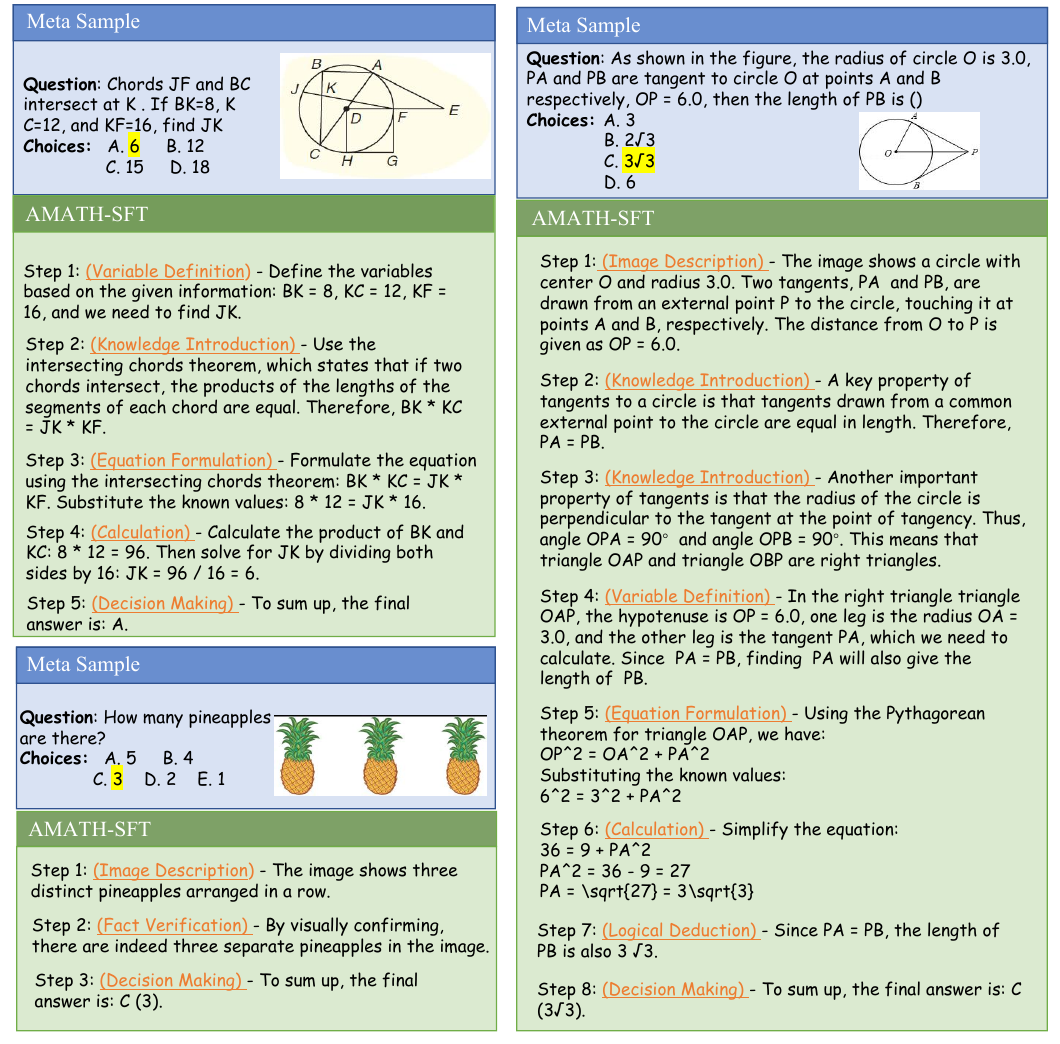} 
        \caption{Case of AMATH-SFT dataset. It includes various mathematical problems such as geometry questions and object counting. Depending on the difficulty of problem, we generate shorter or longer CoTs.}
    \label{fig:dataset}
\end{figure*}

\section{Analysis }
\subsection{What Kind of Challenges Exist in Synthesizing Reasoning Data? }
High cost of manual annotation makes the collection of high-quality reasoning data a bottleneck in developing models toward AGI. In Figure~\ref{fig:dataset}, the high-quality reasoning steps in the AMATH dataset demonstrate the feasibility of data synthesis through intelligent agents. During the synthesis process, we identify three prevalent error types in GPT-4o's outputs: incorrect modeling (including theorem application and logical deduction), flawed assumptions, and visual information misidentification. We analyze 200 error samples generated by GPT-4o using CoT reasoning, revealing a modeling error rate exceeding 60\%, while computational and visual recognition errors accounted for 14.5\% and 21.5\%, respectively. Examples include the erroneous application of the Pythagorean theorem in Figure~\ref{fig:data_engine} Question 1 and the misidentification of Angle 1 and Angle 6 in Problem 3’s diagram. Although these errors can be mitigated through our data engine, extensive data filtering remains necessary, reducing overall data utilization efficiency. Future efforts should prioritize enhancing generative models’ visual reasoning capabilities—specifically, improving visual information perception, accurate modeling based on observations, and proper knowledge utilization. Additionally, test-time scaling strategies such as multiple sampling may improve the data generation success rates.

\subsection{What Kind of Capabilities Do MLLM Need in Reasoning? }
Building upon the set of atomic capabilities illustrated in Figure~\ref{fig:pie_outline}, we calculate our model's utilization rate for each category of steps using Eq~\ref{ability_score}. In Figure~\ref{fig:score_outline}, after employing AtomThink, the success rate of subsequent reasoning steps improves in nearly all the atomic capabilities. Notably, enhancement is more pronounced for some intermediate steps, such as Logical Deduction (with improvements of 32.1\% and 38.8\% on 7B and 11B models, respectively) and Image Description (showing gains of 57.2\% and 21.4\%, respectively). However, the method provides limited benefits for spatial reasoning(0\% in 11B model). Moreover, results reveal that as the given historical steps approach the beginning of reasoning chain (e.g. Image Description and Data Extraction), prediction error rate continuously increases. This error accumulation effect prompts us to focus on the quality of reasoning in initial stages. In future work, we can mitigate the rate of error accumulation by adjusting data ratios and designing sampling strategies.

\subsection{What Kind of Information Do PRM Focus on? }
In Table~\ref{tab:main}, we find that even the problem heavily relies on visual dominant inputs, the highest performance is achieved by using a language PRM. This demonstrates the importance of language model for multimodal reasoning capabilities.  \textcolor{improvered}{To further validate this, we trained both text-only and multimodal PRMs on Llama3.2-Vision-11B using reasoning steps from AMATH and the PRM800K dataset. As shown in Table~\ref{tab:mathverse_prm}, MM-PRM demonstrates only a marginal 0.4\% improvement over its text-only counterpart. Notably, on the Vision Only subset of MathVerse, Text-PRM even slightly outperforms MM-PRM (22.8\% vs. 22.3\%). This indicates that current PRMs still mainly rely on language models to obtain supervisory signals from textual information.}  Exploring how to leverage multimodal features to correct the reasoning process will be a direction we need to investigate.

\subsection{\textcolor{improvered}{Case Study}}

\begin{figure*}[th]
    \centering
    \includegraphics[width=1\textwidth]{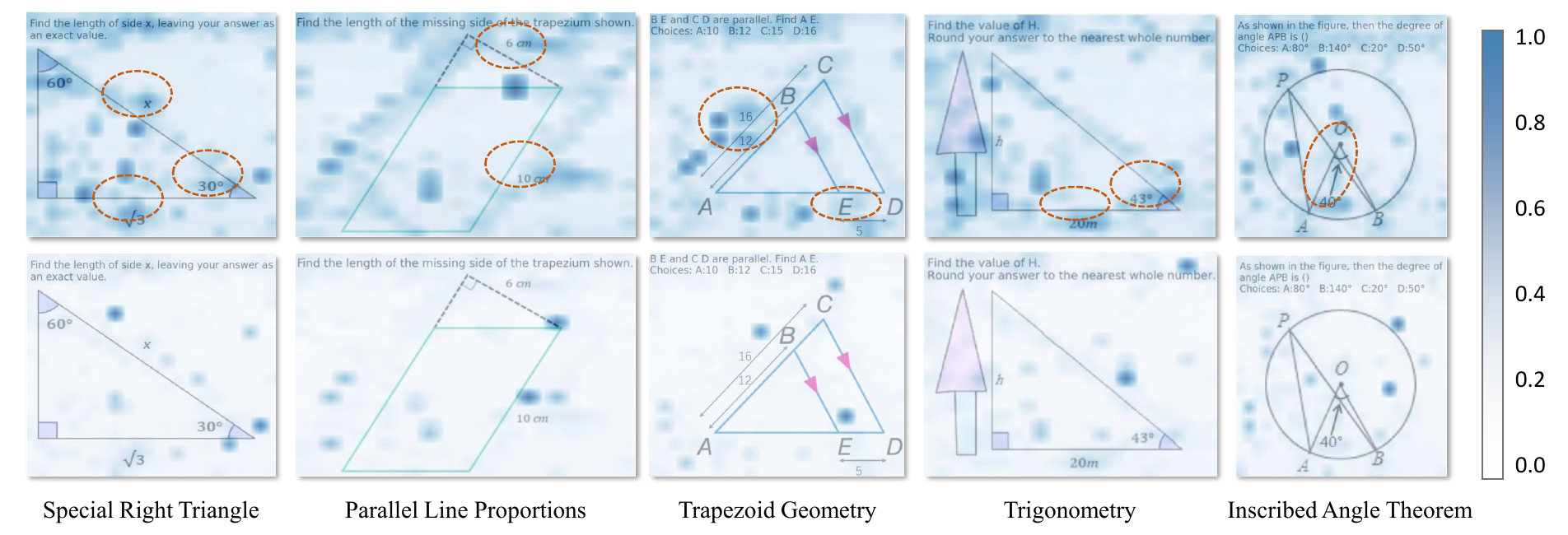} 
    \caption{\textcolor{improvered}{Attention map visualization comparison between AtomThink-LLaVA (top row) and LLaVA-1.5-7B (bottom row) on five representative geometry problem types. Red dashed circles highlight key geometric features (e.g., angles, sides, numerical labels) that are critical for problem-solving. Our model demonstrates more focused attention on these problem-relevant visual regions.}}
    \label{fig:case_study}
\end{figure*}

\noindent\textbf{\textcolor{improvered}{Attention Maps.} } 
\textcolor{improvered}{To further illustrate the enhancement in visual perception capabilities, we compare attention maps between AtomThink-LLaVA and LLaVA-1.5-7B in Figure~\ref{fig:case_study}. The inputs are image-only, with question text incorporated into the images themselves. Compared to LLaVA-1.5-7B, AtomThink-LLaVA demonstrates significantly improved attention allocation across all problem types. Specifically, our model places substantially more attention on text regions while effectively focusing on key geometric features such as angle markers, side lengths, and special symbols. For instance, in the Parallel Line Proportions problem, AtomThink-LLaVA correctly attends to the critical measurement "10 cm" and "6 cm", which are essential for applying the proportional relationship. Similarly, in the Trigonometry case, our model focuses on both the angle marker "$43\angle$" and the base length "20m", demonstrating proper identification of the parameters needed for trigonometric calculation. In contrast, LLaVA-1.5-7B shows more dispersed attention patterns with substantial focus on task-irrelevant background regions. However, AtomThink-LLaVA still attends to many irrelevant regions, indicating noisy attention patterns that require further optimization in future work. For example, in the Inscribed Angle Theorem case, both models allocate non-trivial attention to the background area outside the circle, suggesting room for improvement in filtering out distracting visual information.}

\begin{figure*}[th]
    \centering
    \includegraphics[width=0.99\textwidth]{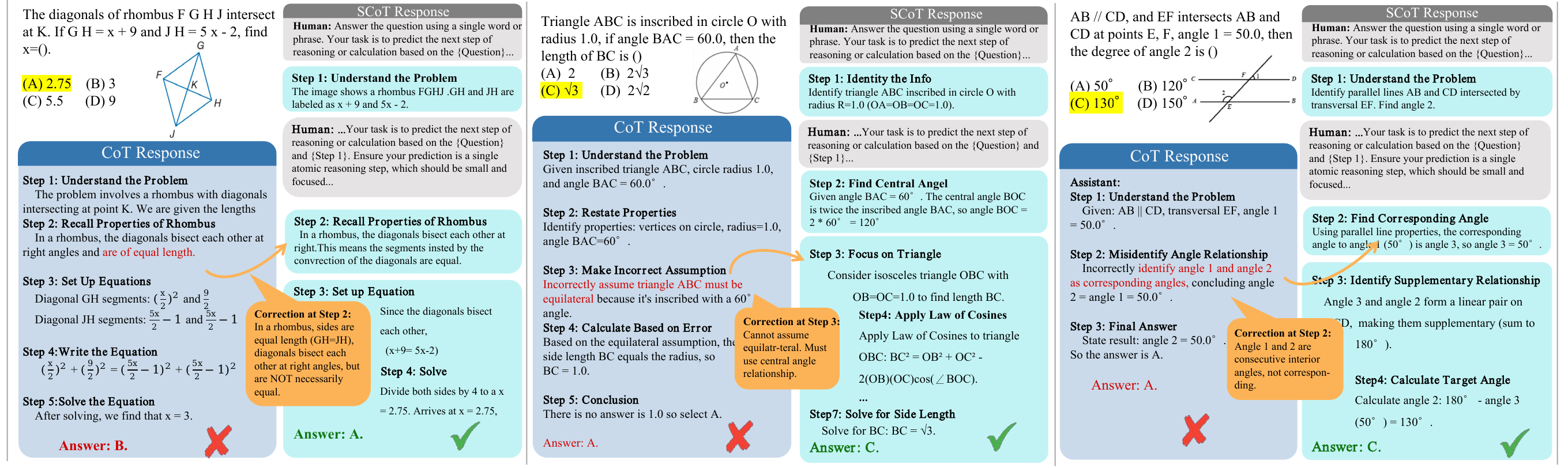} 
        \caption{\textcolor{improvered}{Error correction examples of SCoT. AtomThink-LlamaV corrects misunderstandings of mathematical theorems (left figure), errors in assumptions (middle) and geometric angle identification (right figure).}}
    \label{fig:case_study2}
\end{figure*}

\noindent\textbf{\textcolor{improvered}{Error Correction Examples. }} 
\textcolor{improvered}{To demonstrate the tangible benefits of atomic step reasoning, Figure~\ref{fig:case_study2} presents case studies where SCoT corrects errors in vanilla CoT: (1) Theorem misapplication: preventing property confusion in the rhombus problem by separating recall from calculation, (2) Erroneous assumptions: identifying unjustified equilateral assumptions in the triangle problem through explicit decomposition, and (3) Visual misinterpretation: correctly distinguishing supplementary from corresponding angles in the parallel lines problem. Atomic reasoning mitigates these errors by mandating explicit articulation of each step and continuous grounding in the input, while the diverse fine-tuning dataset provides broad reasoning knowledge that prevents hidden assumptions and hallucinations.}

\subsection{\textcolor{improvered}{Exploration in Reinforcement Learning}}
\textcolor{improvered}{The emergence of reinforcement learning has provided a new paradigm for the evolution of reasoning capabilities. In Figure~\ref{fig:grpo}, we verify the effectiveness of AMATH dataset within GRPO~\cite{guo2025deepseek} framework. Using the setup from R1V~\cite{yu25r1vision}, we extract a 3K-sized subset of AMATH, with a scale comparable to R1V-CLEVR, and conduct experiments on Qwen2-VL-2B~\cite{yang2024qwen2}. The reward function consists of format reward and accuracy reward. Figure~\ref{fig:grpo} shows the changes in response length and accuracy rewards during training. Although accuracy improves in the early stages of training (first 30 steps), it subsequently stabilized around 30\% without converging. This may be due to the higher complexity of AMATH leading to overly sparse training rewards. Additionally, neither experimental group exhibits a CoT length growth phenomenon observed in DeepSeek-R1~\cite{guo2025deepseek}. Since AtomThink's step-level search significantly increases computational resource requirements based on dialogue turns, we leave process reward reinforcement learning for future research.}

%% file: tables/main_exp.tex
\begin{table*}[h!]
\centering
\caption{Comparison of accuracy with state-of-the-art models on \sout{four  benchmarks}\textcolor{improvered}{mathematical task}. Our AtomThink achieves consistent improvement across models of varying scales and surpasses baselines on all four benchmarks. Specially, AtomThink-LlamaV, with 11B parameters, surpasses GPT-4V by 8.5\% on MathVista \textcolor{improvered}{and by 3.2\% on WeMath}. The baseline models (*) are post-trained by LLaVA100K VQA.}
\resizebox{\textwidth}{!}{ 
\begin{tabular}{
    c 
    c 
    |
    c 
    c 
    c 
    c
    |
    c
}
\toprule
\textbf{Model} & \textbf{Inference} & \textbf{MathVista} & \textbf{MathVerse} & \textbf{MathVision} & \textbf{\textcolor{improvered}{WeMath}} & \textbf{\textcolor{improvered}{Avg}}  \\
\midrule
Random Choice & - & 17.9 & 12.4 & 7.2 & - & -\\
Human & - & 70.9 & - & 68.8 & - & - \\
OpenAI o1 & CoT &73.9 & 57.0 & 60.3 & - & -\\
Claude 3.5 Sonnet &CoT&67.7&-&38.0 & - & -\\
GPT-4o & CoT  &63.8 & 50.2 & 30.4 & 50.6 & 48.8 \\
GPT-4V & CoT  &49.9& 54.4 & 24.0 & 51.4 & 44.9 \\
LLaVA-NeXT-34B & Direct & 46.5  & 23.8  & -& - & -\\
InternLM-XComposer2 & Direct & 57.6 & 16.5 & 14.5& 30.9 & 29.9 \\
Qwen-VL-Plus & Direct &43.3  &11.8 & 10.7 & - & -\\
LLaVA-1.5-13B &Direct&27.6 & 15.6&11.2& - & -\\
\midrule
LLaVA1.5-7B*& Direct & 27.3&10.0&9.3& 13.0 & 14.9 \\
AtomThink-LLaVA & SCoT & 
  29.4{\textcolor{improvegreen}{\small\ (+1.9)}} & 
  14.4{\textcolor{improvegreen}{\small\ (+4.4)}} & 
 12.7{ \textcolor{improvegreen}{\small\ (+3.4)}} &
  34.5 \textcolor{improvegreen}{{\small\ (+21.5)}}&
 22.8 { \textcolor{improvegreen}{\small\ (+7.9)}} \\
AtomThink-LLaVA & SCoT w/ PRM & 
  32.1\textcolor{improvegreen}{{\small\ (+4.8)}} & 
  14.6\textcolor{improvegreen}{{\small\ (+4.6)}} & 
  12.3\textcolor{improvegreen}{{\small\ (+3.0)}} & 
  44.2 \textcolor{improvegreen}{{\small\ (+31.2)}}&
  25.8\textcolor{improvegreen}{{\small\ (+11.9)}} \\
\midrule
Llama3.2-Vision-11B*& Direct &47.5&23.3&13.8 & 16.6 & 25.3 \\
AtomThink-LlamaV & SCoT & 
  57.1\textcolor{improvegreen}{{\small\ (+9.6)}} & 
  31.5\textcolor{improvegreen}{{\small\ (+8.2)}} & 
  18.2\textcolor{improvegreen}{{\small\ (+4.4)}} &
  51.9 \textcolor{improvegreen}{{\small\ (+35.3)}} &
  39.7 \textcolor{improvegreen}{{\small\ (+14.4)}} \\
AtomThink-LlamaV & SCoT w/ PRM & 
  58.4\textcolor{improvegreen}{{\small\ (+10.9)}} & 
  33.5\textcolor{improvegreen}{{\small\ (+10.2)}} & 
 21.0\textcolor{improvegreen}{{\small\ (+7.2)}} &
 54.6\textcolor{improvegreen}{{\small\ (+38.0)}} &
 41.9 \textcolor{improvegreen}{{\small\ (+16.6)}} \\

\bottomrule
\end{tabular}
}
\label{tab:main}
\end{table*}

%% file: tables/general_exp.tex
\begin{table*}[h!]
\centering
\caption{\textcolor{improvered}{Comparison of accuracy with state-of-the-art models on general and scientific reasoning benchmarks. general ability benchmarks include DocVQA, ChartQA, and TextVQA; Scientific reasoning task includes ScienceQA, AI2D, MMMU, and HLE. Even with fine-tuning only on mathematical reasoning data, AtomThink-LlamaV exhibits impressive generalization ability, with gains of 25.3\% on ChartQA and 20.2\% on ScienceQA.}}
\resizebox{\textwidth}{!}{ 
\begin{tabular}{
    c 
    c 
    |
    c 
    c 
    c 
    |
    c
    c
    c
    c
}
\toprule
&  & \multicolumn{3}{c|}{\textbf{\textcolor{improvered}{General Tasks}}} &  \multicolumn{4}{c|}{\textbf{\textcolor{improvered}{Scientific Tasks}}} \\
\midrule
\textbf{\textcolor{improvered}{Model}} & \textbf{\textcolor{improvered}{Inference}} & \textbf{\textcolor{improvered}{DocVQA}} & \textbf{\textcolor{improvered}{ChartQA}} & \textbf{\textcolor{improvered}{TextVQA}} & \textbf{\textcolor{improvered}{ScienceQA}} & \textbf{\textcolor{improvered}{AI2D}} & \textbf{MMMU} & \textbf{HLE}  \\
\midrule
Human & - & 98.1 & - & 86.0 & 90.2 & 95.2 & 88.6 & - \\
OpenAI o1 & CoT & - & - & - & - & - & 78.2 & 8.8\\
Claude 3.5 Sonnet & CoT & 95.2 & 90.8 & 74.1 & 81.2 & 80.2 & 68.3 & 4.8\\
GPT-4o & CoT & 92.8 & 92.8 & 77.4 & 88.2 & 86.3 & 69.1 & 3.1\\
GPT-4V & CoT & 88.4 & 78.5 & 78.0 & - & 78.6 & 56.8 & -\\
\midrule
Llama3.2-Vision-11B* & Direct & 62.4 & 59.4 & 68.0 & 65.6 & 62.4 & 42.7 & 4.0 \\
AtomThink-LlamaV & SCoT & 
  66.6\textcolor{improvegreen}{{\small\ (+4.4)}} & 
  78.1\textcolor{improvegreen}{{\small\ (+18.7)}} & 
  72.8\textcolor{improvegreen}{{\small\ (+4.8)}} &
  85.9\textcolor{improvegreen}{{\small\ (+20.3)}} &
  65.6\textcolor{improvegreen}{{\small\ (+3.2)}} &
  47.6\textcolor{improvegreen}{{\small\ (+4.9)}} &
  5.4\textcolor{improvegreen}{{\small\ (+1.4)}} \\
AtomThink-LlamaV & SCoT w/ PRM & 
  68.8\textcolor{improvegreen}{{\small\ (+6.4)}} & 
  84.7\textcolor{improvegreen}{{\small\ (+25.3)}} & 
  80.2\textcolor{improvegreen}{{\small\ (+12.2)}} &
  85.8\textcolor{improvegreen}{{\small\ (+20.2)}} &
  73.4\textcolor{improvegreen}{{\small\ (+11.0)}} &
  48.0\textcolor{improvegreen}{{\small\ (+5.3)}} &
  4.5\textcolor{improvegreen}{{\small\ (+0.5)}} \\

\bottomrule
\end{tabular}
}
\label{tab:main2}
\end{table*}

%% file: tables/llavacot_and_data_scaling.tex
\begin{table*}[th]
\centering
\begin{minipage}{0.63\textwidth} 
\caption{Comparison with LLaVA-CoT. We not only improve inference accuracy by 3.6\%, but also decrease the data and test-time resource requirement.}
\centering
\resizebox{\columnwidth}{!}{  
\begin{tabular}{l | c c c c}
\toprule
\textbf{Method} & MathVista  & \textbf{Dataset Scale} & \textbf{Tokens} & \textbf{Inference Time}\\
\midrule
LLamaV-o1 & 54.4 & 174k (86k in SCI) & - & - \\
LLaVA-CoT & 54.8 & 100k (28.9k in SCI) & 1322.2 & 57.2 \\
AtomThink-LlamaV & 57.1\textcolor{improvegreen}{{\small\ (+2.3)}} & 20k\textcolor{improvegreen}{{\small\ (-80\%)}} & 161.5\textcolor{improvegreen}{{\small\ (-87.8\%)}} & 8.4\textcolor{improvegreen}{{\small\ (-85.3\%)}} \\
AtomThink w/ PRM & 58.4\textcolor{improvegreen}{{\small\ (+3.6)}} & 20k\textcolor{improvegreen}{{\small\ (-80\%)}} & 734.7\textcolor{improvegreen}{{\small\ (-44.4\%)}} & 38.1\textcolor{improvegreen}{{\small\ (-33.4\%)}} \\
\bottomrule
\end{tabular}
}
\label{tab:llavacot}
\end{minipage}
\hfill
\begin{minipage}{0.35\textwidth} 
\caption{\textcolor{improvered}{Dataset scaling experiments using AtomThink-LLaVA and MathVision-mini. For the ablation experiments on LLaVA-Instruct, we keep AMATH-SFT at a scale of 124k. For AMATH-SFT ablations, we keep LLaVA-Instruct at 100k.}}
\centering
\resizebox{\columnwidth}{!}{ 
\begin{tabular}{l | c c c c c}
\toprule
\textbf{LLaVA-Instruct}& 0k & 50k & 100k & 200k & 400k \\
\textbf{Accuracy} & 8.81 & 10.98 & 12.45 & 12.24 & 12.50 \\
\midrule
\textbf{AMATH-SFT} & 0k & 10k & 30k & 60k & 124k \\
\textbf{Accuracy} & 9.28 & 9.67 & 9.33 & 11.33 & 12.45 \\
\bottomrule
\end{tabular}
}
\label{tab:data_scaling}
\end{minipage}
\end{table*}


%% file: tables/test_time_scaling.tex
\begin{table}[t]
\caption{AtomThink-LlamaV performance improvement of MathVision-mini with test-time scaling. We employ Best-of-N and PRM to select the optimal step among N candidates.}
\centering
\resizebox{\columnwidth}{!}{  
\begin{tabular}{c c c}
\toprule
\textbf{Test Time Scaling} & \textbf{Ouput Tokens} & \textbf{Accuracy}\\
\midrule
BS Candidate=0        & 2.3  & 13.9  \\
BS Candidate=1   & 231.9  & 18 \\
BS Candidate=2        & 518.6  & 18.3 \\
BS Candidate=3       & 822.3  &23.3 \\
\bottomrule
\end{tabular}
}
\label{tab:tt_scaling}
\end{table}

%% file: tables/more_models.tex
\begin{table*}[t]
\caption{Performance of more models on mathematical reasoning tasks. Our AtomThink-LLaVA-Llama3 outperforms the baseline in all sub-tasks across two benchmarks, achieving an average improvement of 14.2\%. In AtomThink-EMOVA, it improves the MathVerse accuracy by 5.4\%.}
\centering
\resizebox{\textwidth}{!}{ 
\begin{tabular}{
    c 
    c 
    |
    c 
    c 
    c 
    |
    c 
    c 
    c 
    c 
    c 
    c 
}
\toprule
& & \multicolumn{3}{c|}{\textbf{MathVista}} & \multicolumn{6}{c}{\textbf{MathVerse}} \\
\midrule
\textbf{Model} & \textbf{Inference} & \textbf{General} & \textbf{Math} & \textbf{Total} & \textbf{TL} & \textbf{TD} & \textbf{VI} & \textbf{VD} & \textbf{VO} & \textbf{Total} \\

\midrule
LLaVA-Llama3-8B & Direct & 34.1 & 25.6 & 29.5 & 16.0 & 19.3 & 16.4 & 13.1 & 15.0 & 15.9\\
w/. Formatted & CoT & 30.2 & 22.9 & 26.3 & 14.3 & 18.4 & 15.7 & 10.0 & 7.7 &13.2 \\
AtomThink-Llama3& Direct & 34.4 & 27.2 & 30.5 & 16.0 & 19.3 & 16.2 & 13.1 & 15.0 & 15.9\\
AtomThink-Llama3& SCoT & \textbf{36.9} & \textbf{37.0} & \textbf{36.6} & \textbf{22.2} & \textbf{26.6} & \textbf{24.1} & \textbf{20.9} & \textbf{17.9} & \textbf{22.4}\\
AtomThink& SCoT w./ PRM & \textbf{36.5} & \textbf{41.3} & \textbf{39.1} & \textbf{36.1} & \textbf{42.4} &\textbf{30.0}  &\textbf{36.8}  & \textbf{28.6 }&\textbf{34.7} \\
\midrule
EMOVA-8B & Direct & 52.4 & 51.1 & 51.7 & 34.4 & 39.0 & 33.4 & 30.1 & 23.5 & 32.1 \\
w/. Formatted & CoT & 30.9 & 31.3 & 31.1 & 26.5 & 36.5 & 25.3 & 20.4 & 19.8 & 25.7 \\
AtomThink-EMOVA& Direct & 53.9 & 52.4 & 53.1 & 33.6 & 39.0 & 33.8 & 28.0 & 24.4 & 31.8\\
AtomThink-EMOVA& SCoT & 48.7 & \textbf{54.4} & \textbf{51.8} & \textbf{36.5} & \textbf{42.4} & \textbf{34.1} & \textbf{32.9} & \textbf{29.7} & \textbf{35.1} \\
AtomThink-EMOVA& SCoT w./ PRM & 48.9 & \textbf{57.0} & \textbf{53.3} &\textbf{42.1}& \textbf{51.5}&\textbf{39.0} &\textbf{36.7} &\textbf{33.1} & \textbf{40.5} \\
\bottomrule
\end{tabular}
}
\label{tab:app_main}
\end{table*}

%% file: tables/search_method.tex
\begin{table}[ht]
\caption{Ablation study on Path-wise and step-wise search. The results show that both Best-of-N-Min(BoN-Min) and Beam Search exhibit consistent performance improvements.}
\centering
\resizebox{\columnwidth}{!}{ 
\begin{tabular}{
    l 
    l 
    |
    c 
    c
    c 
    c 
    c 
}
\toprule
\textbf{Model} & \textbf{Method} & \textbf{MathVista-M} & \textbf{MathVista-G} & \textbf{MathVerse} \\
\midrule
EMOVA-200k & Direct & 51.1 & 52.4 & 33.3 & \\
\midrule
\multicolumn{1}{l}{\multirow{2}*{AtomThink}} & Direct & 52.4 & 53.9 & 35.7 & \\
 & SCoT & 54.2 & 46.7 &38.0 &\\
\midrule
\multicolumn{1}{l}{\multirow{4}*{w/. Path-wise}}  & Majority Voting & 48.8 & 49.4 & 39.0 & \\
 & BoN-Last & 51.2 & 46.8 & 41.3 & \\
 & BoN-Avg & 58.7 & 40.5 & 38.7 \\
 & BoN-Min & 53.7 & 53.2 & 40.0 \\
 \midrule
\multicolumn{1}{l}{\multirow{3}*{w/. Step-wise}} & Greedy & 46.3 & 45.6 & 38.3 & \\
 & Beam Search & 57.1 & 53.2 & 45.3 \\
& \textcolor{improvered}{MCTS} & 57.8 & 52.2 & 47.6 \\

\bottomrule
\end{tabular}
}
\label{tab:search}
\end{table}

%% file: tables/ablation.tex
\begin{table*}[t]
\caption{Ablation study in MathVista.}
\centering
\resizebox{\textwidth}{!}{ 
\begin{tabular}{
    c 
    c 
    |
    c 
    |
    c 
    c 
    c 
    c 
    c 
    c 
    c 
    c 
}
\toprule
\textbf{Model} & \textbf{Inference} & \textbf{Total} & \textbf{Statistical} & \textbf{Scientific} & \textbf{Geometry} & \textbf{Arithmetic} & \textbf{Algebraic} & \textbf{Numeric} & \textbf{Logical} \\
\midrule
LLaVA1.5-7B& Direct & 27.3 & 24.2 & 45.1 & 25.1 & 22.8 & 28.4 & 17.3 & 13.5 \\
w/o Formatted & CoT & 27.0 & 20.5 & 33.7 & 31.4 & 23.8 & 30.9 & 16.3 & 15.8 \\
w/o Formatted & SCoT & 27.5 & 20.3 & 44.3 & 36.8 & 19.6 & 33.5 & 18.1 & 8.1 \\
w/o Formatted & SCoT w./ PRM & 28.8 & 21.3 & 45.9 & 38.1 & 20.9 & 35.9 & 18.1 & 8.1 \\
AtomThink & CoT & 28.8 & 24.2 & 41.8 & 26.4 & 30.3 & 26.7 & 22.2 & 16.2 \\
AtomThink & SCoT & 29.4 & 25.3 & 42.6 & 28.9 & 30.9 & 25.6 & 25.0 & 16.2 \\
AtomThink & SCoT w./ PRM & 32.1 & 25.3 & 42.6 & 36.4 & 30.9 & 35.9 & 22.9 & 13.5 \\
\midrule
Llama3.2-Vision-11B & Direct & 47.5 & 61.3 & 61.4 & 44.2 & 43.8 & 43.4 & 31.9 & 16.2 \\
w/o Formatted & CoT & 48.4 & 62.8 & 63.1 & 44.4 & 43.0 & 43.7 & 32.6 & 8.1 \\
w/o Formatted & SCoT & 44.3 & 55.2 & 59.3 & 46.7 & 39.9 & 41.3 & 30.2 & 0.0 \\
w/o Formatted & SCoT w./ PRM & 48.9 & 58.5 & 59.8 & 49.8 & 39.9 & 49.8 & 35.4 & 10.8 \\
AtomThink & CoT & 50.4 & 67.8 & 59.8 & 48.1 & 43.9 & 47.7 & 25.7 & 13.5 \\
AtomThink & SCoT & 57.1 & 69.1 & 62.3 & 55.7 & 50.1 & 55.9 & 38.9 & 16.2 \\
AtomThink & SCoT w./ PRM & 58.4 & 68.8 & 63.9 & 61.5 & 51.6 & 60.5 & 39.6 & 5.4 \\
\bottomrule
\end{tabular}
}
\label{tab:ablation}
\end{table*}

%% file: tables/prm.tex
\begin{table}[th]
\caption{ \textcolor{improvered}{Performance comparison on MathVerse benchmark with text based and multimodal PRMs. We split MathVerse with 5 subset by vision dependency, include Text Lite (TL), Text Dominant (TD), Vision Intensive (VI), Vision Dominant (VD), Vision Only (VO). The models used include: 
Baseline: Llama3.2-Vision-11B;
AtomThink: AtomThink-LlamaV;
Text: Text based PRM;
MM: Multimodal PRM;
}}
\centering
\resizebox{\columnwidth}{!}{ 
\begin{tabular}{
    l 
    l 
    |
    c 
    c 
    c 
    c 
    c 
    c 
}
\toprule
\textbf{\textcolor{improvered}{Model}} & \textbf{\textcolor{improvered}{PRM}} & \textbf{\textcolor{improvered}{TD}} & \textbf{\textcolor{improvered}{TL}} & \textbf{\textcolor{improvered}{VI}} & \textbf{\textcolor{improvered}{VD}} & \textbf{\textcolor{improvered}{VO}} & \textbf{\textcolor{improvered}{Total}} \\
\midrule
Baseline & Direct & 36.0 & 24.3 & 23.0 & 22.7 & 18.6 & 24.9 \\
Baseline & CoT & 31.6 & 23.2 & 24.8 & 21.8 & 17.9 & 23.9 \\
\midrule
AtomThink & MM & 38.5 & 33.5 & 32.4 & 29.8 & 22.3 & 31.3 \\
AtomThink & Text & 38.5 & 33.2 & 31.2 & 28.6 & 22.8 & 30.9 \\
AtomThink & Text(Qwen) & 45.9 & 35.9 & 34.0 & 33.0 & 24.6 & 34.7 \\

\bottomrule
\end{tabular}
}
\label{tab:mathverse_prm}
\end{table}

%% file: sec/5_limitation.tex
\section{Limitation}
Due to the high cost of manual annotation, we only \textcolor{improvered}{conduct} a sampled quality check on the dataset. Although \textcolor{improvered}{the error rate is} low, our dataset still suffers from issues such as redundant expressions and unclear granularity in atomic step segmentation \textcolor{improvered}{with} a gap \textcolor{improvered}{remaining compared to} step-level gold-standard annotation. This may be attributed to insufficient instruction-following \textcolor{improvered}{capabilities} of existing data generation models.

Additionally, to balance data production costs and reasoning quality, we \textcolor{improvered}{do} not employ other open-source models as data engines\textcolor{improvered}{which introduces} a performance bottleneck in AtomThink's training. \textcolor{improvered}{Since} our data sources come from publicly available databases or training sets, they may overlap with the pre-training data of the latest models \textcolor{improvered}{and potentially lead} to generalizability issues.

Finally, \textcolor{improvered}{the Self-structured Chain-of-Thought in this method} is learned through \textcolor{improvered}{a supervised fine-tuning strategy}, which places high demands on both data scale and quality. Due to the challenges of parameter tuning and limited computational resources, it has not been extended to  more complex training \textcolor{improvered}{methods such as} Reinforcement Learning from Human Feedback (RLHF), which may constrain the upper bounds of performance and generalization.

%% file: sec/6_conclusion.tex
\section{Conclusion}
To mitigate overthinking and encourage structured output, we  propose a self-structured chain of thought method \textcolor{improvered}{that ensures} reasoning efficiency while adaptively generating a diverse taxonomy of atomic steps. Subsequently, we \textcolor{improvered}{introduce} AtomThink, a comprehensive deep reasoning framework that encompasses data engineering, model training, inference, and evaluation. The experimental results demonstrate that our method consistently \textcolor{improvered}{improves} \textcolor{improvered}{the model's} diverse behaviors \textcolor{improvered}{at} test time \textcolor{improvered}{and enhances} reasoning performance across various multimodal benchmarks. This work paves the way for developing \textcolor{improvered}{more generalizable}  slow-thinking models and offers insights for understanding \textcolor{improvered}{multimodal} reasoning patterns.

%% file: sec/X_suppl.tex

\appendix  
\subsection{Implementation Details}
\label{appendix: imple}

\subsubsection{Policy Models}
\label{appendix: policy models}
In this section, we provide more implementation details for baseline models and our framework. Firstly, we post-train LLaVA1.5-7B and Llama3.2-Vision-11B using AMATH-SFT and a sub-sampled dataset of LLaVA665K, containing 100k samples. During this process, the weights of LLM, projector and vision encoder are fully fine-tuned. Specifically, we utilize the Llama-factory~\cite{zheng2024llamafactory} framework to train the models and the hyperparameters are listed in Table\ref{tab:parameters}. For LLaVA-Llama3~\cite{liu2024visual}, we choose the pre-trained ViT-L/14 of CLIP~\cite{radford2021learning} as the vision encoder and Llama3-8B~\cite{dubey2024llama} as our LLM. To align visual features with the LLM, we incorporates a Multi-Layer Perceptron (MLP) as a projector between the visual encoder and the language model. For EMOVA-8B~\cite{chen2024emova}, we use the original setting of EMOVA that uses InternViT-6B~\cite{chen2024internvl} and LLaMA-3.1-8B~\cite{dubey2024llama}. The C-Abstractor~\cite{cha2024honeybee} with two ResBlocks is adopted as the projector. The training of LLaVA-Llama-3-8B follows a structured two-stage process~\cite{liu2024visual}. In our experiment, we only load its weights from pre-training stage and deploy supervised fine-tuning. During SFT, the training data comprises the LLaVA-Instruct-665k, a 46k subset of PRM800k and our AMATH-SFT dataset. The weights of language model and MLP projector are unfreezed. The model undergoes an epoch of training with a reduced learning rate of 2e-5 and batch size of 128. To create AtomThink-EMOVA, we post-train EMOVA using AMATH-SFT and a sub-sampled dataset of EMOVA-SFT-4m, containing 200k samples. During this process, the weights of the LLM and the C-Abstractor projector are updated. EMOVA is fine-tuned for 1 epoch with a batch size of 128 and a learning rate of 2e-6.

\begin{table*}[ht]
    \caption{Comparison of Parameters for post-training LLaVA1.5-7B and Llama-3.2-Vision-11B.}
    \centering
    \begin{tabular}{l|cccc}
    \toprule
    \textbf{Parameter} & \textbf{LLaVA1.5-7B} & \textbf{Llama3.2-V-11B} & \textbf{LLaVA-Llama-3-8B} & \textbf{EMOVA-8B} \\ 
    \midrule
    Learning Rate & 2e-6 & 2e-6 & 2e-5 & 2e-6\\ 
    Epochs & 1 & 1  & 1 & 1\\ 
    Batch Size & 128 & 128 & 128 & 128\\ 
    Context Length & 4096 & 4096 & 4096 & 4096 \\ 
    Seed & 42 & 42 & 42 & 42 \\ 
    Precision & FP16 & BF16 & FP16 & BF16 \\ 
    GPU & 32*32G V100& 8*80G A800 & 32*32G V100& 8*80G A800\\ 
    FSDP & True & True & True & True\\ 
    DeepSpeed & Zero3 & Zero3 & Zero3 & Zero3\\
    \bottomrule
    \end{tabular}
    \label{tab:parameters}
\end{table*}

\begin{figure*}[ht]
    \centering
    \includegraphics[width=0.9\textwidth]{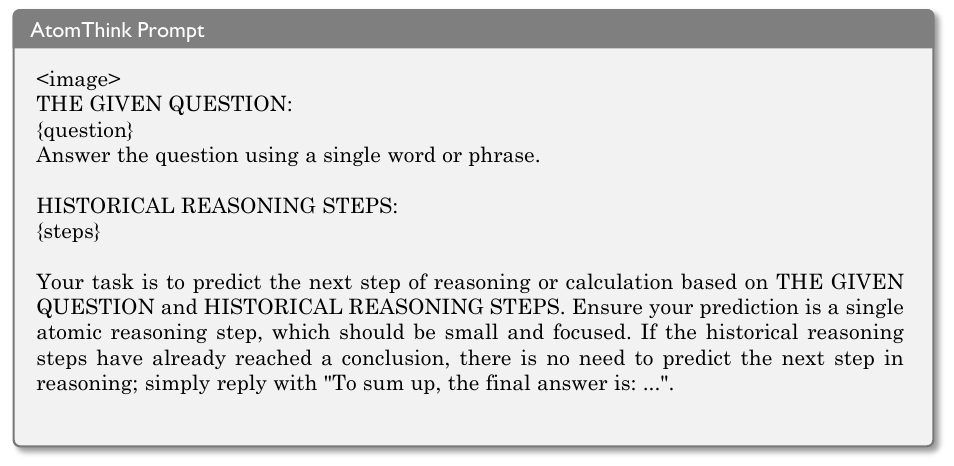} 
        \caption{AtomThink template for generating Self-structured CoT. The model takes an image and a question as input, generating an atomic step at each iteration. These steps are then concatenated into the historical reasoning steps, which are fed into model for the next round of reasoning.}

    \label{fig:prompts_atomthink}
\end{figure*}

\subsubsection{PRM Setting} 
\label{appendix: prm}
We initially use a open-sourced large language model (Qwen2.5-Math-PRM-7B~\cite{prmlessons}) to introduce textual process supervision. Results of our main experiments in Table~\ref{tab:main} demonstrate that AtomThink can be seamlessly integrated with such external models in a plug-and-play manner.  Furthermore, in Table~\ref{tab:app_main} we also fine-tune a PRM based on Math-psa-7B~\cite{wang2024openr} model as our foundational architectures. Math-psa-7B is a text-based process supervision model trained using datasets such as PRM800K~\cite{lightman2023let}, Math-Shepherd~\cite{Math-shepherd} and MATH-APS~\cite{wang2024openr}. Low-Rank Adaptation (LoRA) is applied to fine-tune with the following parameters: rank (r) of 8, alpha scaling factor of 32, dropout rate of 0.1, and targeting the q and v projectors. Training is conducted over one epoch with a batch size of 256 and a learning rate of 1e-5. We sample a 20k-instance training set from PRM800K and combine it with the AMATH-PRM dataset, which is derived from multimodal CoT annotations, to serve as our fine-tuning data. All the samples include question, historical steps, and current step, with each current step being assigned a label of either correct or incorrect. We designate "\textbackslash n\textbackslash n\textbackslash n\textbackslash n\textbackslash n" as the step separator and return the conditional probability of the current step being correct.

\subsection{Prompts Design}
\label{appendix: prompts}
\begin{figure*}[t]
    \centering
    \includegraphics[width=0.9\textwidth]{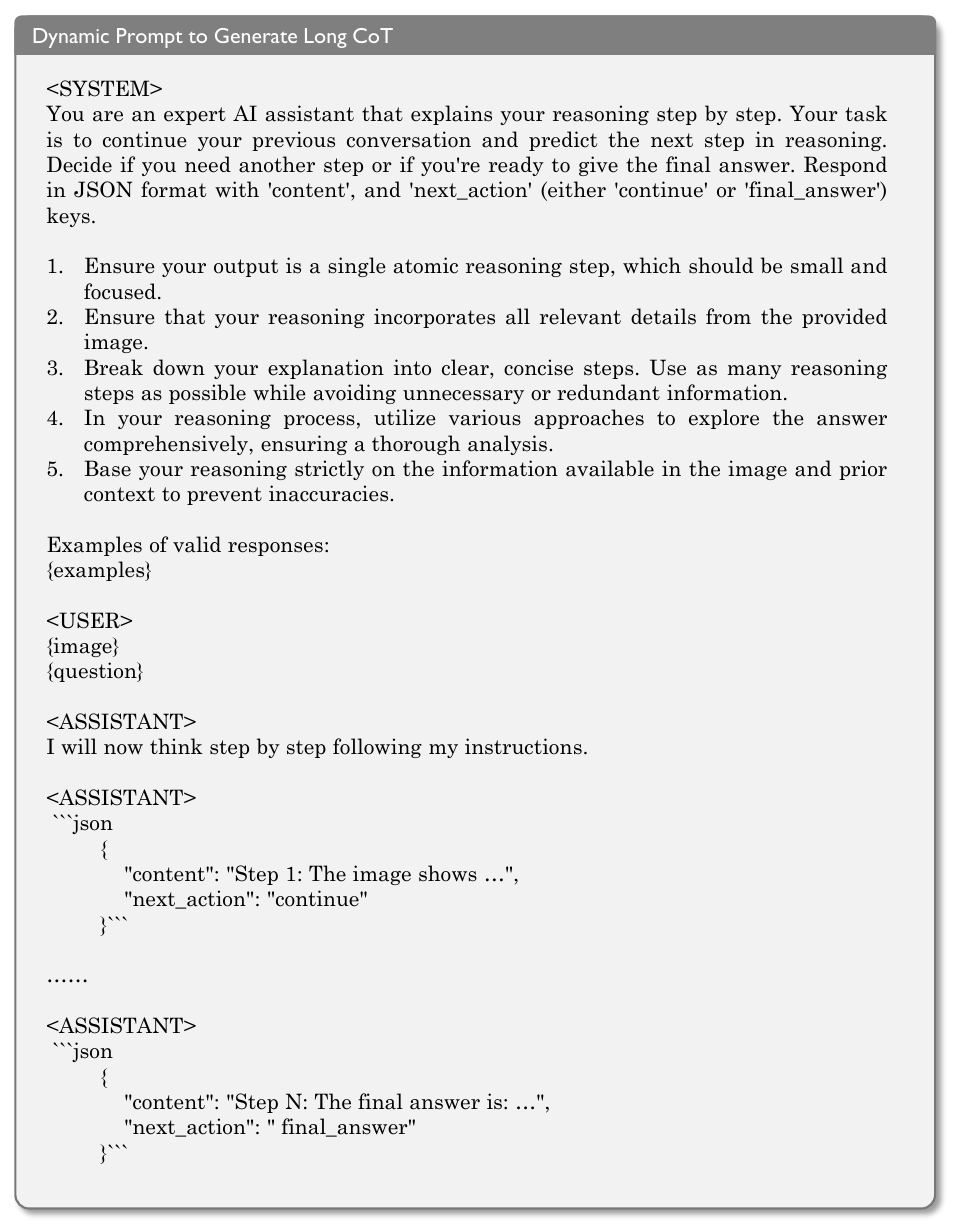} 
        \caption{Dynamic prompt for long CoT generation. Inspired by previous work, we designed a dynamic prompt template that generates reasoning steps for each iteration. It effectively identifies the input visual information to generate detailed image captions and fine-grained atomic steps.}

    \label{fig:prompts_dynamic}
\end{figure*}

\begin{figure*}[t]
    \centering
    \includegraphics[width=\textwidth]{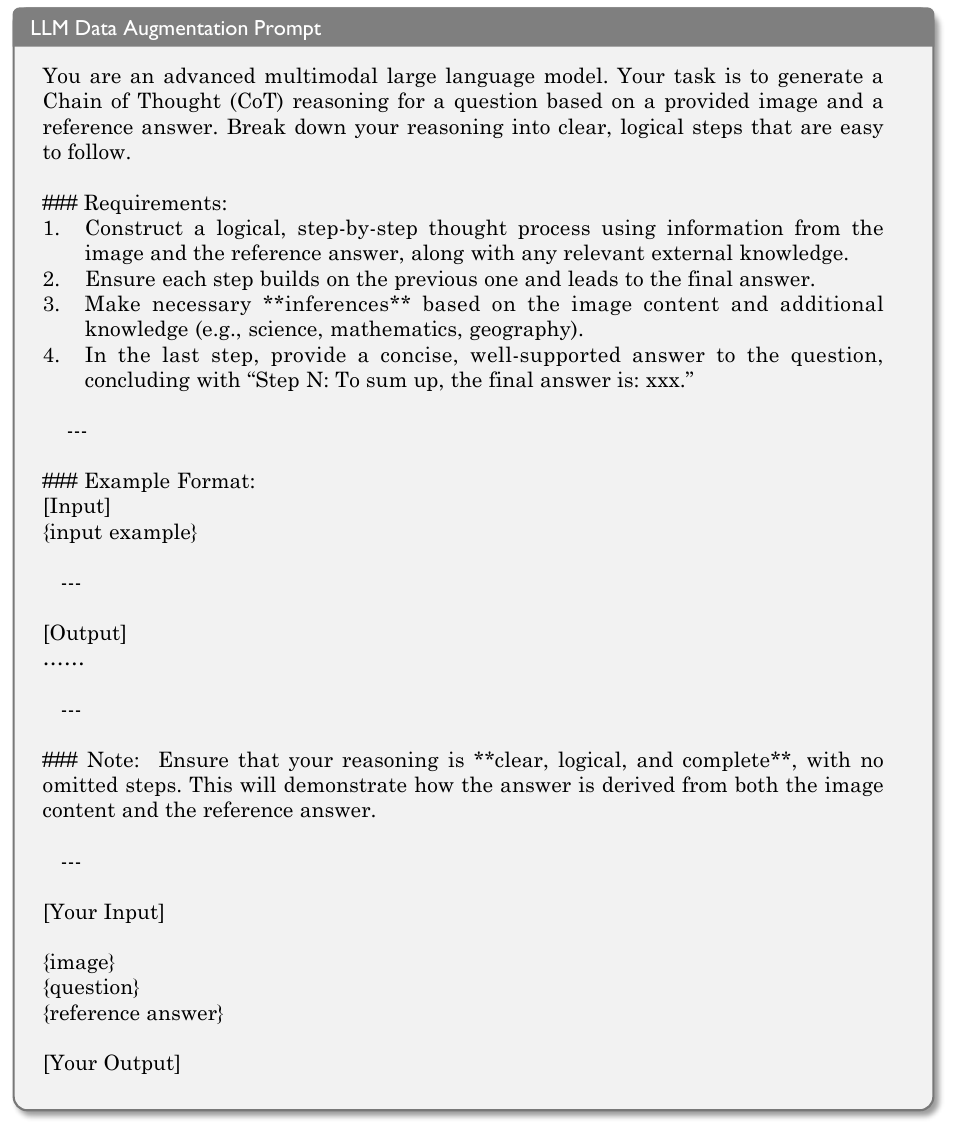} 
        \caption{Prompt for short answer augmentation. Using the current math VQA dataset, which already includes short answers and CoTs, we apply this template to enhance and generate detail atomic steps.}
\label{fig:prompts_cot_aug}
\end{figure*}

\begin{figure*}[t]
    \centering
    \includegraphics[width=\textwidth]{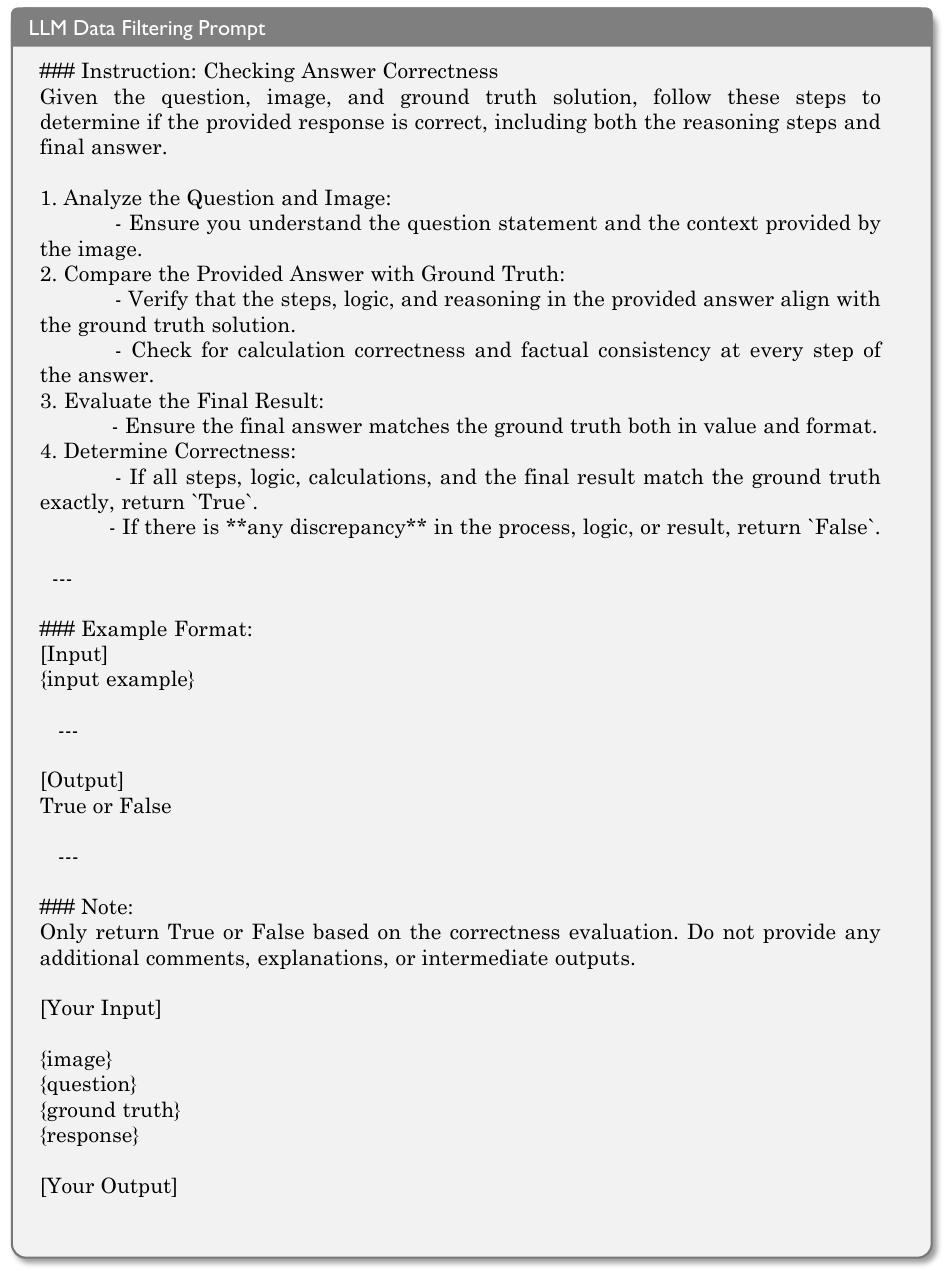} 
        \caption{Prompt for filtering wrong CoT. Due to the quality gap between the reasoning steps generated by the AI assistant and human annotations, we employ this template to double-check. It filters out samples with incorrect answers and reasoning processes.}

    \label{fig:prompts_data_filtering}
\end{figure*}

\begin{figure*}[t]
    \centering
    \includegraphics[width=0.85\textwidth]{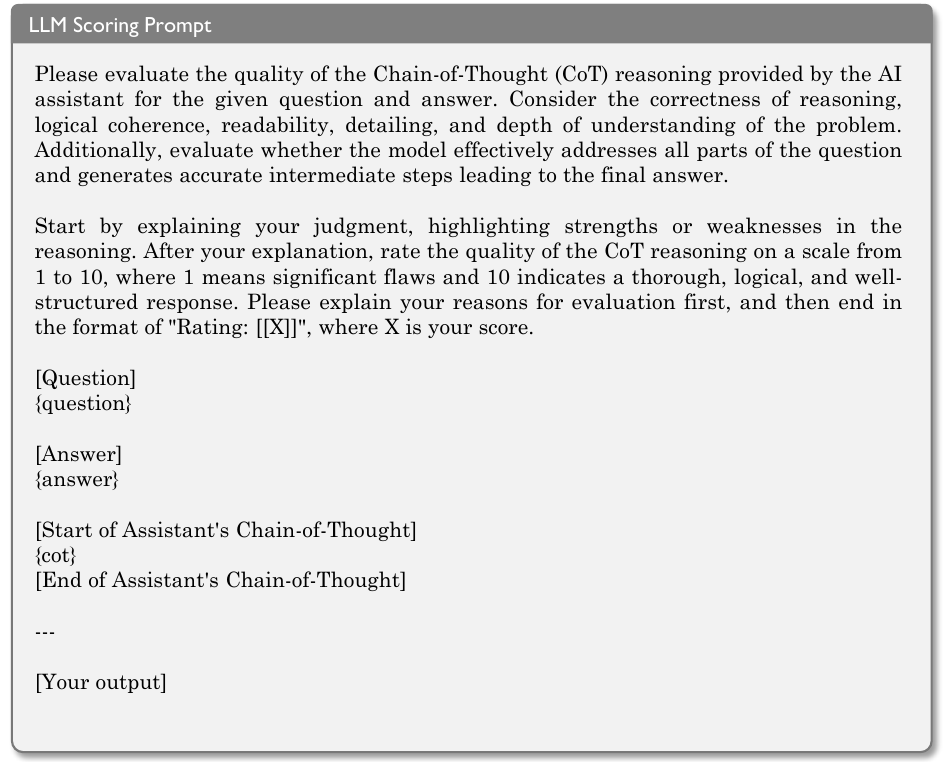} 
        \caption{Prompt for GPT scoring. We use this template and GPT-4o to quantitatively evaluate the quality of the generated data. The results show that our AMATH data outperforms human annotations in terms of AI preference scores.}

    \label{fig:prompts_scoring}
\end{figure*}

In this section, we present the prompt used in self-structured CoT~\ref{fig:prompts_atomthink} and multimodal CoT annotation engine. Prompts in data engine include: long CoT generation (Figure \ref{fig:prompts_dynamic}), short CoT augmentation (Figure \ref{fig:prompts_cot_aug}), data filtering (Figure \ref{fig:prompts_data_filtering}), and quality scoring (Figure \ref{fig:prompts_scoring}).


\begin{figure*}[t]
    \centering
    \includegraphics[width=0.95\textwidth]{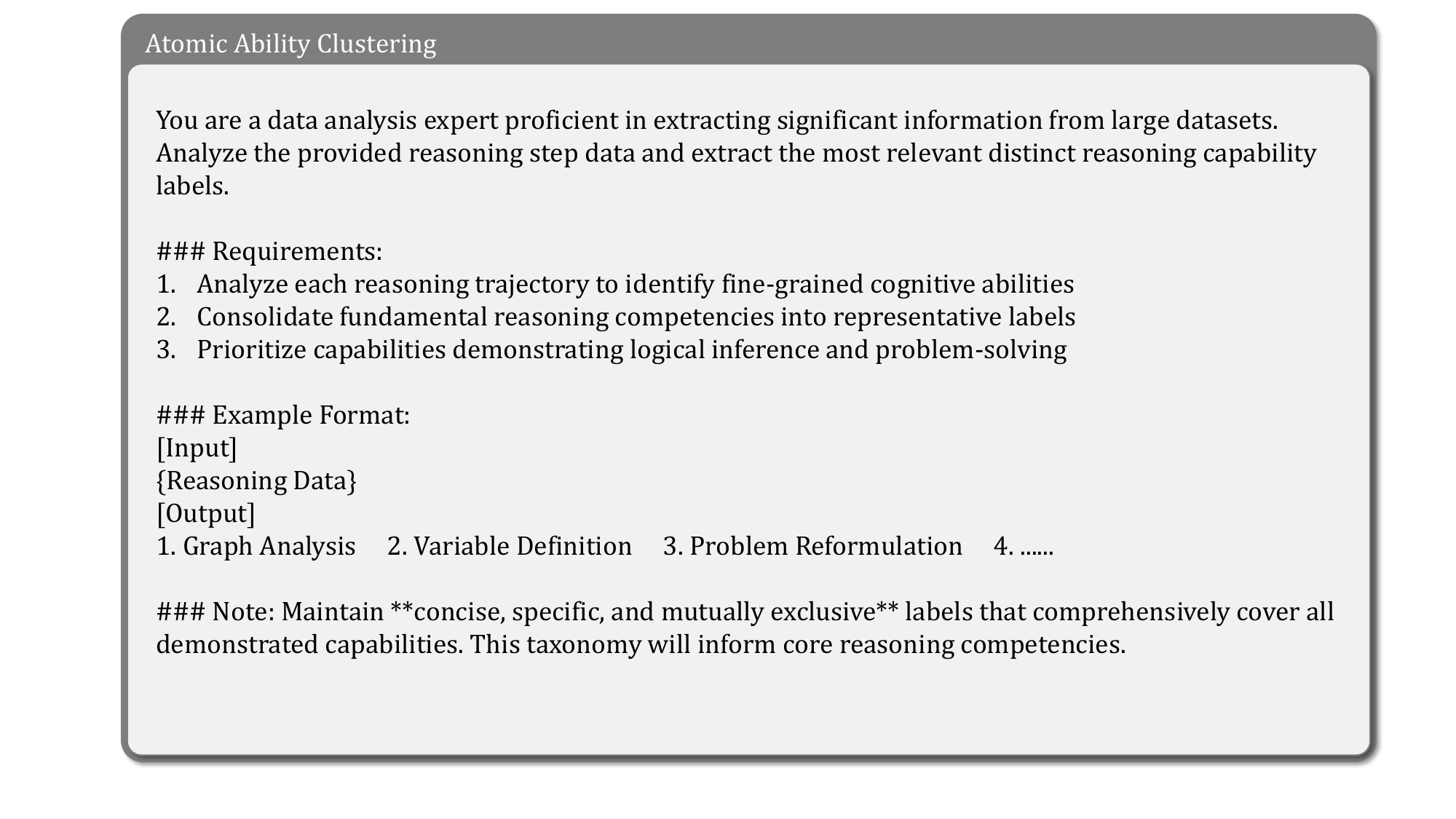} 
        \caption{Prompt for clustering the reasoning behaviors with Kimi1.5.}

    \label{fig:atomic_ability}
\end{figure*}
